\documentclass{article}
\usepackage[margin=2.5cm]{geometry}

\usepackage[utf8]{inputenc}
\usepackage[T1]{fontenc}
\usepackage{graphicx} 
\usepackage{physics}
\usepackage{amssymb}  
\usepackage{mathtools} 

\usepackage{amsthm}
\theoremstyle{plain}

\theoremstyle{remark}
\newtheorem{remark}{Remark}
\newtheorem{problem}{Problem}

\usepackage{empheq}
\usepackage{verbatim}
\usepackage{soulutf8}

\usepackage[
    hyperfootnotes=false,
    colorlinks=true,
    linkcolor=blue,
    urlcolor=blue,
    citecolor=blue,
    anchorcolor=blue,
    pagebackref=false,
    unicode=true]{hyperref}

\usepackage{multirow}
\usepackage{readarray}

\usepackage{physics}
\DeclareDocumentCommand\pmqty{ l m }{\begin{pmatrix}#1 #2 \end{pmatrix}}
\DeclareDocumentCommand\He{ l m }{\mathrm{He}\qty#1{#2}}
\DeclareDocumentCommand\Ve{ l m }{\mathbf{Ve}\pqty#1{#2}}
\DeclareDocumentCommand\Co{ l m }{\mathbf{Co}\pqty#1{#2}}
\DeclareDocumentCommand\Gr{ l m }{\mathbf{Gr}\pqty#1{#2}}
\DeclareDocumentCommand\ls{ l m }{\mathrm{Ls}\qty#1{#2}}

\NewDocumentCommand\NotCov{
    s       
    t\tilde 
    t\hat   
    t\bar   
    m       
    g     
    o       
    o       
    }{%
    {%
        \IfBooleanTF{#2}{\tilde#5}{%
        \IfBooleanTF{#3}{\hat  #5}{%
        \IfBooleanTF{#4}{\bar  #5}{%
                               #5}}%
        }\IfNoValueF{#7}{^{#7}}_{\IfNoValueF{#6}{#6}\IfNoValueF{#8}{\IfNoValueF{#6}{,}#8}}%
    }%
}
\NewDocumentCommand{\Sigmawz}{o}{\NotCov\Sigma{wz}[#1]}

\title{\LARGE \bf
Tendon-based modelling, estimation and control for a simulated high-DoF anthropomorphic hand model%
\thanks{Project no. TKP2021-NKTA-66 has been implemented with the support provided by the Ministry of Technology and Industry of Hungary from the National Research, Development and Innovation (NRDI) Fund, financed under the TKP2021-NKTA funding scheme. This research was also supported by the NRDI Office through the OTKA grant no. PD-145902. P.P. gratefully acknowledges the research grant no. BO/00427/25 supported by the János Bolyai Research Scholarship of the Hungarian Academy of Sciences.}
}

\author{Péter Polcz$^{1}$, Katalin Schäffer$^{1,2}$, and Miklós Koller$^{1}$}

\date{\small
$^{1}$Faculty of Information Technology and Bionics, Pázmány Péter Catholic University, Budapest, Hungary, \newline (e-mail: {\tt polcz.peter|schaffer.katalin|koller.miklos@itk.ppke.hu})
\\[4pt]
$^{2}$Department of Aerospace and Mechanical Engineering, University of Notre Dame, Notre Dame, IN 46556, USA (e-mail: {\tt kschaff2@nd.edu})
}

\usepackage{setspace}
\begin{document}

\maketitle
\thispagestyle{empty}
\pagestyle{empty}

\begin{abstract}
Tendon-driven anthropomorphic robotic hands often lack direct joint angle sensing, as the integration of joint encoders can compromise mechanical compactness and dexterity. This paper presents a computational method for estimating joint positions from measured tendon displacements and tensions. An efficient kinematic modeling framework for anthropomorphic hands is first introduced based on the Denavit-Hartenberg convention. Using a simplified tendon model, a system of nonlinear equations relating tendon states to joint positions is derived and solved via a nonlinear optimization approach. The estimated joint angles are then employed for closed-loop control through a Jacobian-based proportional-integral (PI) controller augmented with a feedforward term, enabling gesture tracking without direct joint sensing. The effectiveness and limitations of the proposed estimation and control framework are demonstrated in the MuJoCo simulation environment using the Anatomically Correct Biomechatronic Hand, featuring five degrees of freedom for each long finger and six degrees of freedom for the thumb.
\end{abstract}


\begin{figure}
    \centering
    \includegraphics[width=\columnwidth]{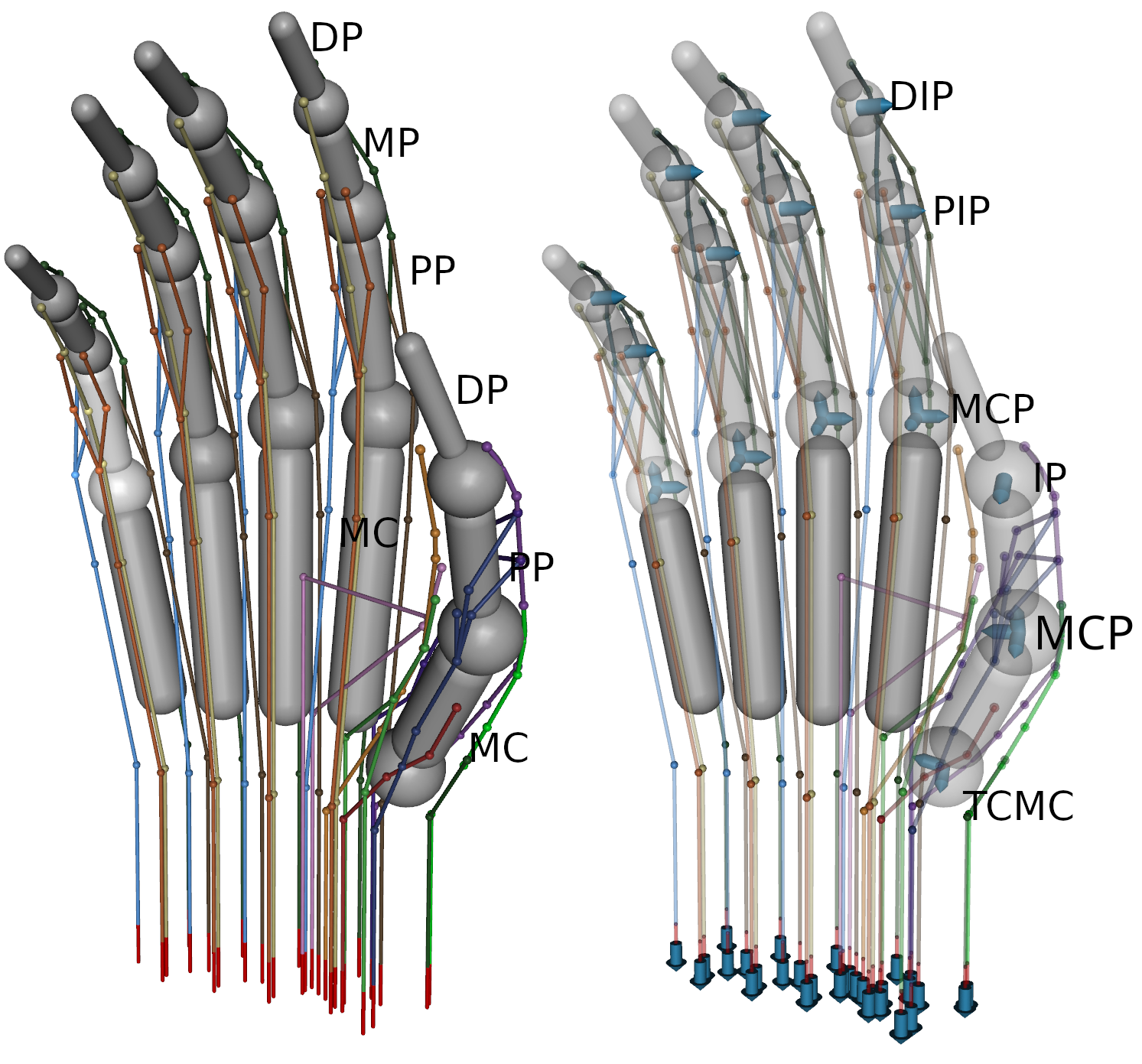}
    \caption{Bone structure of the ACB Hand at rest position. Abbreviations of the bones are given in the left figure, whereas, the right figure illustrate the joints with their names, and their degrees of freedom (i.e., the axes of free rotations). This illustration is a slightly modified version of that appeared in \cite{2024_Polcz.etal}.}
    \label{fig:Bone_Structure}

\end{figure}

\section{Introduction}

In robotics, it is still a challenge to develop anthropomorphic hand models that can imitate the dexterity of the human hand. A commonly used approach to achieve better biological accuracy is to mimic the entire biological structure of the hand, including the tendon-based actuation mechanism of the hand
\nocite{2016_Xu.Todorov,2018_MohdFaudzi.etal,2022_Konda.etal,2019_Zhang.etal,2021_Min.Yi,2022_Shafer.Deshpande,2021_Tian.etal,2019_Tasi.etal}
\cite{2016_Xu.Todorov}--\cite{2019_Tasi.etal}.

In particular, the Anatomically Correct Biomechatronic (ACB) Hand developed by Tasi et\,al. \cite{2019_Tasi.etal} precisely follows the shapes and structures of a human hand to mimic human hand's dexterity as much as possible.
In such robotics systems the bones of the fingers can be displaced along more degrees of freedom compared to the segments of a conventional robotic manipulator.
Differently from \cite{2021_Min.Yi}--\cite{2021_Tian.etal}, the metacarpals of long fingers in \cite{2019_Tasi.etal}, for example, can rotate about all the three axes (Pitch, Yaw, and Roll).
For such models, even kinematic description is difficult using the Denavit-Hartenberg (DH) convention, see, e.g., \cite{2022_Huczala.etal,2022_Biswal.Parida}.

%
Needless to say that the precise knowledge of the joint posture is essential for control algorithms \cite{2017_Niehues.etal,2019_Lange.etal,2022_Sievers.etal}.
However, as encoder installed to the (possibly multiple DoF) joints would compromise the mobility of the hand, the joint angles are generally not measured but estimated, e.g., from the tendon excursions using feedforward neural networks \cite{2017_Niehues} or through motion capture \cite{2020_Esmatloo.Deshpande}. As demonstrated in \cite{2019_Kim.etal}, joint coupling also allows exact gesture computation from tendon excursions, but it significantly reduces the manipulators functionality compared to a human hand.

This manuscript builds upon our previous conference publication \cite{2024_Polcz.etal}, whose primary contribution was an analytical joint posture estimation method for the Anatomically Correct Biomechatronic (ACB) Hand based on solving a system of nonlinear equations derived from tendon displacements and tensions. That work also introduced a systematic Denavit-Hartenberg-based kinematic description for finger-type manipulators composed of irregularly connected three-degree-of-freedom joints, and demonstrated an iterative, model-based estimation and control framework as an interpretable alternative to data-driven approaches. While the analytical nature of the method enables explicit cause-effect analysis and facilitated tasks such as in-silico tendon placement optimization, the conference paper also identified limitations in tracking certain gestures and transient motions. The present manuscript extends this prior work by incorporating a feedforward term into the Jacobian-based PI control scheme, resulting in improved posture tracking accuracy and reduced settling time. A comparative analysis of the controller performance with and without the feedforward term is provided to quantify the benefits of the proposed enhancement.


\section{Compute joint angles and excursion of connected tendon segments}
\subsection{Bone structure and its kinematic model}
In this paper, we follow the bone structure developed by Tasi et\,al. \cite{2019_Tasi.etal}, where the bones are connected by 1, 2, or 3 DoF joints.
The flexion-extension is described by Pitch rotation, adduction-abduction is the Yaw rotation, and the pronation-supination is given by Roll rotation.
The metacarpophalangeal (MCP) joints of ACB Hand has 3 DoF, allowing the proximal phalanges (PP) to be displaced in both Pitch, Yaw and Roll directions.
The thumb carpometacarpal (TCMC) joint has 2 DoF, thus the metacarpal (MC) of the thumb can freely rotate about both Pitch and Yaw axes.
The proximal interphalangeal (PIP), distal interphalangeal (DIP) joints, and the interphalangeal (IP) of the thumb have a free hinge motion only about the Pitch axis.
The bone structure is illustrated in Figure \ref{fig:Bone_Structure}.

The irregular alignment of the cartilages causes two bones to deflect and twist relative to each other (even in the rest position). These irregularities are modeled by offset angles. The deflection is described Pitch and Yaw offsets, whereas, the twist is given by the Roll offset.

In our geometric model, the bone structure is built such that the alignment of two bones can be described by six angles: three fixed offset angles and the rotations about the three free joints.
One possible way to construct a DH table for such robotic manipulators with $n$ links and 3-DoF joints is given in Table \ref{DH:general}, where $(X_\mathrm{o}, Y_\mathrm{o}, Z_\mathrm{o})$ is the coordinate of the first joint and $L_k$ is the length of the $k$th link. This convention allows the removal of the appropriate pair of $(3j,3j+1)$ rows when a Yaw rotation is missing, e.g., when Joint 1 Yaw vanishes, rows $(3,4)$ are negligible. Using the presented convention, the DH tables of a long finger and the thumb are given in Tables \ref{DH:Fx} and \ref{DH:T1}, respectively.

\begin{remark}
    In the case of the long fingers, PIP and DIP Yaw offsets are zero, therefore, further rows can be removed from the DH tables. Similar situation occurs with the thumb, as it does not have a specific Yaw offset at the MCP and IP joints.
\end{remark}

\begin{table}
    \renewcommand{\arraystretch}{1.05}
    \centering
    \begin{small}
    \begin{tabular}{|@{~}c@{~}|@{~~}l@{~~}l@{~~}|@{~~}l@{~~}l@{~~}|}
    \hline
    $i$     &  $d$  &  $\vartheta$  &  $a$   &  $\alpha$ \\
    \hline
    \hline
            &  $Z_\mathrm{o}$  &  0       &  $X_\mathrm{o}$  &  $-90^\circ$ \\
            &  $Y_\mathrm{o}$  &  $-90^\circ$   &  0   &  \textit{Joint $1$ Roll Offset} \\
    \hline
    \hline
    1       &  0  &  \textit{Joint $1$ Yaw Offset}   &  0   &  $-90^\circ$ \\
    2       &  0  &  \textit{Joint $1$ Pitch Offset}   &  0   &  \textbf{Joint $1$ Roll} \\
    3       &  0  &  0       &  0   &  $90^\circ$ \\
    \hline
    4       &  0  &  \textbf{Joint $1$ Yaw}       &  0   &  $-90^\circ$ \\
    5       &  0  &  \textbf{Joint $1$ Pitch}       &  $L_1$  &  \textit{Joint $2$ Roll Offset} \\
    6       &  0  &  0       &  0   &  $90^\circ$ \\
    \hline
    \hline
    $\cdots$ & & $\cdots$ & & $\cdots$ \\
    \hline
    $6k\!-\!5$    &  0  &  \textit{Joint $k$ Yaw Offset}   &  0   &  $-90^\circ$ \\
    $6k\!-\!4$  &  0  &  \textit{Joint $k$ Pitch Offset}   &  0   &  \textbf{Joint $k$ Roll} \\
    $6k\!-\!3$  &  0  &  0       &  0   &  $90^\circ$ \\
    \hline
    $6k\!-\!2$  &  0  &  \textbf{Joint $k$ Yaw}       &  0   &  $-90^\circ$ \\
    $6k\!-\!1$  &  0  &  \textbf{Joint $k$ Pitch}       &  $L_k$  &  \textit{Joint $k\!+\!1$ Roll Offset} \\
    $6k$  &  0  &  0       &  0   &  $90^\circ$ \\
    \hline
    \hline
    $\cdots$ & & $\cdots$ & & $\cdots$ \\
    \hline
    \hline
    $6n\!-\!2$  &  0  &  \textbf{Joint $n$ Yaw}       &  0   &  $-90^\circ$ \\
    $6n\!-\!1$  &  0  &  \textbf{Joint $n$ Pitch}       &  $L_n$  &  0 \\
    \hline
    \end{tabular}
    \end{small}
    \caption{DH table for a manipulator with $n$ links connected irregularly through 3-DoF rotary joints}
    \label{DH:general}

    \smallskip

    \begin{small}
    \begin{tabular}{|l|ll|ll|}
    \hline
    $i$     &  $d$  &  $\vartheta$  &  $a$   &  $\alpha$ \\
    \hline
    \hline
            &  $Z_\mathrm{o}$  &  0       &  $X_\mathrm{o}$  &  $-90^\circ$ \\
            &  $Y_\mathrm{o}$  &  $-90^\circ$   &  0   &  \textit{CMC Roll offset} \\
    \hline
    1       &  0  &  \textit{CMC Yaw offset}   &  0   &  $-90^\circ$ \\
    2       &  0  &  \textit{CMC Pitch offset}   &   $L_\textsc{mc}$  &  \textit{MCP Roll offset} \\
    6       &  0  &  0       &  0   &  $90^\circ$ \\
    \hline
    \hline
    7       &  0  &  \textit{MCP Yaw offset}   &  0   &  $-90^\circ$ \\
    8       &  0  &  \textit{MCP Pitch offset}   &  0   &  \textbf{MCP Roll} \\
    9       &  0  &  0       &  0   &  $90^\circ$ \\
    \hline
    10      &  0  &  \textbf{MCP Yaw}   &  0   &  $-90^\circ$ \\
    11      &  0  &  \textbf{MCP Pitch}   &  $L_\textsc{pp}$  &  \textit{PIP Roll offset$^\star$} \\
    \hline
    \hline
    14      &  0  &  \textit{PIP Pitch offset}   &  0   &  0 \\
    \hline
    17      &  0  &  \textbf{PIP Pitch}   &  $L_\textsc{mp}$  &  \textit{DIP Roll offset$^\star$} \\
    \hline
    \hline
    20      &  0  &  \textit{DIP Pitch offset}   &  0   &  0 \\
    \hline
    23      &  0  &  \textbf{DIP Pitch}   &  $L_\textsc{dp}$  &  0 \\
    \hline
    \multicolumn{5}{@{\,}l}{$^\star$PIP and DIP Yaw offsets are missing}
    \end{tabular}
    \end{small}
    \caption{DH table of the 5 DoF long fingers of ACB Hand.}
    \label{DH:Fx}

    \smallskip

    \begin{small}
    \begin{tabular}{|l|ll|ll|}
    \hline
    $i$     &  $d$  &  $\vartheta$  &  $a$   &  $\alpha$ \\
    \hline
    \hline
            &  $Z_\mathrm{o}$  &  0       &  $X_\mathrm{o}$  &  $-90^\circ$ \\
            &  $Y_\mathrm{o}$  &  $-90^\circ$   &  0   &  \textit{TCMC Roll Offset} \\
    \hline
    \hline
    1       &  0  &  \textit{TCMC Yaw Offset}   &  0   &  $-90^\circ$ \\
    2       &  0  &  \textit{TCMC Pitch Offset}   &  0   &  $90^\circ$ \\
    \hline
    4       &  0  &  \textbf{TCMC Yaw}       &  0   &  $-90^\circ$ \\
    5       &  0  &  \textbf{TCMC Pitch}       &  $L_\textsc{mc}$  &  \textit{MCP Roll Offset$^\star$} \\
    \hline
    \hline
    8       &  0  &  \textit{MCP Pitch Offset}   &  0   &  \textbf{MCP Roll} \\
    9       &  0  &  0       &  0   &  $90^\circ$ \\
    \hline
    10      &  0  &  \textbf{MCP Yaw}       &  0   &  $-90^\circ$ \\
    11      &  0  &  \textbf{MCP Pitch}       &  $L_\textsc{pp}$  &  \textit{IP Roll Offset$^\star$} \\
    \hline
    \hline
    14      &  0  &  \textit{IP Pitch Offset}   &  0   &  0 \\
    17      &  0  &  \textbf{IP Pitch}       &  $L_\textsc{dp}$  &  0 \\
    \hline
    \multicolumn{5}{@{\,}l}{$^\star$Yaw offsets are missing at the thumb's MCP and IP joints}
    \end{tabular}
    \end{small}
    \caption{DH table for the 6 DoF thumb finger of ACB Hand.}
    \label{DH:T1}
\end{table}

\subsection{Tendon structure}
For simplicity, we discuss only the index finger, which is manipulated by $n_\mathrm{m} = 5$ muscles, namely, flexor digitorium superficialis (FDS), flexor digitorium profundus (FDP), extensor digitorium communis (EDC), lumbrical (LUM), and ulnar interosseous (UI).
Tendons EDC, UI, and LUM are divided, then, connected again such that they form two lateral bands (LBs), the extensor slip (ES), and the terminal extensor (TE).
We note that this tendon structure corresponds to that presented in \cite{2017_Niehues.etal} and is a simplified version of \cite{2019_Tasi.etal}.

As illustrated in Figure \ref{fig:tendon_structure_F2}, the tendon structure of the index finger is described by a directed acyclic graph having three types of vertices: {starting sites} ($V_\mathrm{m}$), {junctions} ($V_\mathrm{j}$), and {terminal sites} ($V_\mathrm{bn}$).
A starting site is connected to a muscle, whereas, a terminal site is fixed to a bone.
The number of edges starting from a vertex $v$ is called the {out-degree} of the vertex and is denoted by $\diffd_-(v)$.
Similarly, $\diffd_+(v)$ denotes the number of edges arriving to vertex $v$ and is called the {in-degree} if $v$.
A {tendon segment} is a directed edge of the graph, which connects two vertices in four possible ways:
\begin{itemize}
    \item a starting site with a terminal site ($v_\mathrm{m} \to v_\mathrm{bn})$,
    \item a starting site with a junction ($v_\mathrm{m} \to v_\mathrm{j}$),
    \item a junction with a terminal site ($v_\mathrm{j} \to v_\mathrm{bn}$),
    \item a junction with a junction ($v_\mathrm{j} \to v_\mathrm{j}'$),
\end{itemize}
such that every starting ($v_\mathrm{m} \in V_\mathrm{m}$) and terminal sites ($v_\mathrm{bn} \in V_\mathrm{bn}$) and junctions ($v_\mathrm{j} \in V_\mathrm{j}$) satisfy the following
\begin{equation}
    \begin{aligned}
        &\begin{aligned}
            &\diffd_+(v_\mathrm{m}) = 0,~
            &&\diffd_-(v_\mathrm{m}) = 1,~
            \\
            &\diffd_+(v_\mathrm{bn}) = 1,~
            &&\diffd_-(v_\mathrm{bn}) = 0,~
            \\
        \end{aligned}
        \\
        &\min\qty\big(\diffd_+(v_\mathrm{j}),\diffd_-(v_\mathrm{j})) \ge 1,
        \\
        &\max\qty\big(\diffd_+(v_\mathrm{j}),\diffd_-(v_\mathrm{j})) \ge 2.
    \end{aligned}
\end{equation}
In the last condition, we assumed that every junction has at least one degree (in- or out-) greater than 1.
Otherwise, the trivial junction can be omitted such that the tendon segment arriving to the junction can be merged to the segment starting from the junction.

A {root segment} is a tendon segment connected directly to a muscle through a starting site ($v_\mathrm{m}$).
Let $n_\mathrm{m}$ denote the number of root segments and $n_\mathrm{ct}$ the number of tendon segments connected to other tendons. The overall number of tendon segments is $n_\mathrm{s} = n_\mathrm{m} + n_\mathrm{ct}$.
The ordinal numbering of tendon segments is such that the first $n_\mathrm{m}$ segments are roots, followed by $n_\mathrm{ct}$ connected segments.

A {tendon branch} is a sequence of segments connecting a starting site $v_\mathrm{m}$ and a terminal site $v_\mathrm{bn}$, i.e., connecting a muscle with a bone, see, e.g., branch (4) -- (10) -- (15) -- (18) in Figure \ref{fig:tendon_structure_F2}.
The number of tendon branches ($n_\mathrm{b}$) can be counted as follows:
\begin{equation}
    \label{eq:nb_IS_nm_nj_nct}
    n_\mathrm{b} = n_\mathrm{m} + \sum_{i = 1}^{n_\mathrm{j}} \qty\big( d_+(v_{\mathrm{j},i}) - 1 )
    = n_\mathrm{m} - n_\mathrm{j} + \underbrace{\sum_{i = 1}^{n_\mathrm{j}} d_+(v_{\mathrm{j},i})}_{n_\mathrm{ct}}.
\end{equation}
One may observe that the sum of out degrees of junctions correspond to the number of connected tendons $n_\mathrm{ct}$. This fact is already anticipated in \eqref{eq:nb_IS_nm_nj_nct}.
Finally, we may conclude that the number of junctions together with the number of branches are equal to the number of tendon segments, formally:
\begin{equation}
    n_\mathrm{b} + n_\mathrm{j} = n_\mathrm{m} + n_\mathrm{ct} = n_\mathrm{s}.
\end{equation}

The $n_\mathrm{m} = 5$ tendons of the long fingers have $n_\mathrm{b} = 10$ branches (B0--B9) in total:

\begin{equation}
    \begin{aligned}
        \text{FDP:~~} &(1) && \text{[B0]} \\
        \text{FDS:~~} &(2) - J_1\!
        \begin{dcases}
            (6) \\
            (7) \\
        \end{dcases}  && \begin{aligned}\text{[B1]} \\ \text{[B2]}\end{aligned}
        \\
        \text{UI:~~} &(3) - J_2\!
        \begin{dcases}
            (8) - (15) - (18) \\
            (9) - (16) \\
        \end{dcases}  && \begin{aligned}\text{[B3]} \\ \text{[B4]}\end{aligned}
        \\
        \text{EDC:~~} &(4) - J_3\!
        \begin{dcases}
            (10) - (15) - (18) \\
            (11) - (16) \\
            (12) - (17) - (18) \\
        \end{dcases}  && \begin{aligned}\text{[B5]} \\ \text{[B6]} \\ \text{[B7]}\end{aligned}
        \\
        \text{LUM:~~} &(5) - J_4\!
        \begin{dcases}
            (13) - (16) \\
            (14) - (17) - (18) \\
        \end{dcases}  && \begin{aligned}\text{[B8]} \\ \text{[B9]}\end{aligned}
    \end{aligned}
    \label{eq:Branches}
\end{equation}
where the numbers (1)--(18) in parentheses correspond to the 18 tendon segments and labels $J_1$--$J_4$ denote the tendon junctions. Both the segments and junctions are illustrated in Figure \ref{fig:tendon_structure_F2}.


\section{Posture estimation from tendon excursion}

\subsection{Preliminaries}
The index finger has $n_\theta = 5$ free joints (MCP Roll, MCP Yaw, MCP Pitch, PIP Pitch, and DIP Pitch). The angles of free joints are collected in a vector $\theta \in \mathbb{R}^{n_\theta}$. Let $\theta_0 \in \mathbb{R}^{n_\theta}$ denote the angles corresponding to the rest position when all tendons are relaxed.
It is reasonable to assume that an any admissible gesture of the finger can be uniquely determined by a vector of joint angles $\theta$.
However, we do not expect that a tendon excursions pattern uniquely determines a posture and vice-versa.
Nevertheless, a simple kinematic model allows us to compute the length of the piecewise straight tendons passing through some well-defined site points, which are fixed relative to the bones.

\begin{figure}
    \centering
    \includegraphics[width=0.67\columnwidth]{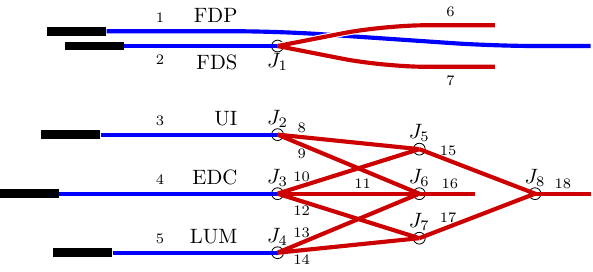}
    \caption{Tendon structure of the index finger. $J_1$--$J_8$ are the labels of the joints, whereas, the numbers correspond to the tendon segments.}
    \label{fig:tendon_structure_F2}

    \bigskip

    \includegraphics[width=0.7\textwidth]{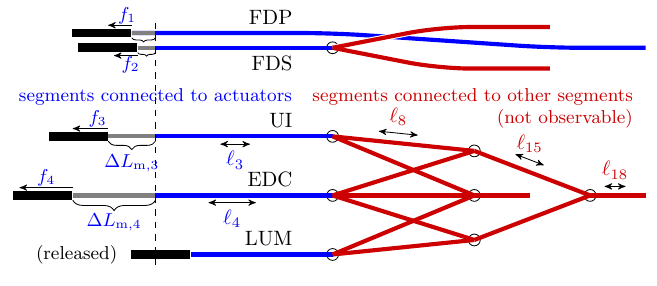}
    \caption{Measured (blue) and unknown variables (red) during gesture estimation.}
    \label{fig:pest_problem}

    \bigskip

    \includegraphics[width=0.7\textwidth]{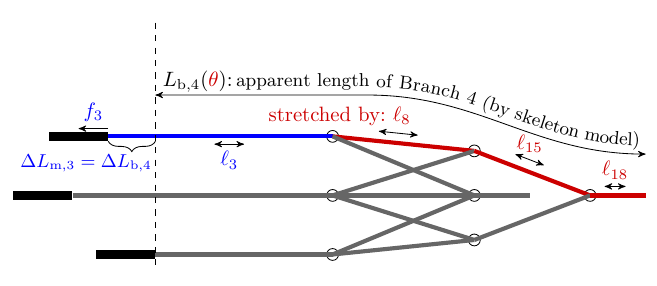}
    \caption{Length of branch calculation during a tendon excursion. Similarly to Figure \ref{fig:pest_problem}, the measured quantities are highlighted in blue, whereas, the unknowns are red.}
    \label{fig:branches}
\end{figure}

Let $L_{\mathrm{s},i}(\theta)$ denote the gesture-dependent length of tendon segment $i \in \qty{1,\dots,n_\mathrm{s}}$ (e.g., segments 1--18 in Figure \ref{fig:tendon_structure_F2}, $n_\mathrm{s} = 18$).
The length of a segment is calculated by the Euclidean distance between its constituent site points.
When a tendon bypasses the joint around the longer arch, the segment is complemented with two auxiliary site points to approximate the curve of the tendon with a piecewise straight tendon structure.
$L_{\mathrm{s}0,i} = L_{\mathrm{s},i}(\theta_0)$ denotes the length of the $i$th segment in the equilibrium position.
The excursion of the $i$th segment is denoted by $\ell_{\mathrm{s},i}(\theta)$.
The acting tension along the $i$th segment is $f_{\mathrm{s},i} = k_i \, \ell_{\mathrm{s},i}$, where $k_i = \frac{E \cdot A}{L_{\mathrm{s}0,i}}$ is the spring (or stiffness) coefficient, $E$ is Young's (tensile) elastic modulus, and $A$ is the cross section area of the tendon.
Then,
$$
    L_\mathrm{s}(\theta)
    = \pmqty{L_\mathrm{m}(\theta) \\ L_\mathrm{ct}(\theta)}
    = \pmqty{L_{\mathrm{s},1}(\theta) \\ \dots \\ L_{\mathrm{s},n_\mathrm{m}}(\theta) \\\hline L_{\mathrm{s},n_\mathrm{m}+1}(\theta) \\ \dots \\ L_{\mathrm{s},n_\mathrm{m} + n_\mathrm{ct}}(\theta)},
$$
similarily
$$
    \ell_\mathrm{s}(\theta)
    = \pmqty{\ell_\mathrm{m}(\theta) \\ \ell_\mathrm{ct}(\theta)}
    \text{, ~~ or~~ }
    f_\mathrm{s} = \pmqty{ f_\mathrm{m} \\ f_\mathrm{ct} }.
$$
Let $\Delta L_{\mathrm{m},i}$ denote the length of stretched tendon segment coiled on the motor shaft, $i \in \qty{1,\dots,n_\mathrm{m}}$.
When $\Delta L_{\mathrm{m},i}$ is positive, tendon $i$ is coiled up on the shaft, whereas a negative $\Delta L_{\mathrm{m},i}$ means that tendon is coiled down relative to the rest position.

\subsection[Excursion from tendon structure model]{Excursion from tendon structure model}
In the knowledge of the joint angles $\theta \in \mathbb{R}^{n_\theta}$, we are able to approximate the length of each tendon segment with the assumption that the site points and junctions are fixed relative to the bones of the finger.
Let ${L_\mathrm{s}} : \mathbb{R}^{n_\theta} \to \mathbb{R}^{n_\mathrm{s}}$ denote this mapping, such that ${L_{\mathrm{s},i}}(\theta)$ denotes the gesture-dependent length of the $i$th segment, $i \in \qty{1,\dots,n_\mathrm{s}}$.
During the model description, the junctions were positioned relatively to the bones in the rest position of the fingers represented by angles $\theta_0$, therefore, the length of tendon segments ${L_{\mathrm{s}0}} = {L_{\mathrm{s}}(\theta_0)}$ in the rest position well approximates the actual lengths of tendons of the physical robotic hand prosthesis.

The length of tendon branches ${L_\mathrm{b}({\color{black} \theta})}$ in an arbitrary gesture ($\theta$) can be computed by the sum of the appropriate tendon segments as follows:
\begin{equation}
    {L_\mathrm{b}({\color{black} \theta})} = C_{\mathrm{b},\mathrm{s}} \, {L_\mathrm{s}({\color{black} \theta})},
\end{equation}
in particular
\begin{equation}
    {L_{\mathrm{b}0}} = C_{\mathrm{b},\mathrm{s}} \, {L_{\mathrm{s}0}} = C_{\mathrm{b},\mathrm{s}} \, {L_\mathrm{s}({\color{black} \theta_0})},
\end{equation}
where $C_{\mathrm{b},\mathrm{s}} \in \mathbb{R}^{{n_\mathrm{b}}\times{n_\mathrm{s}}}$ is a sparse matrix, in which the $(i,j)$th element is $1$ if the $i$th branch contains the $j$th segment and $0$ otherwise. The actual value of $C_{\mathrm{b},\mathrm{s}}$ for the long fingers is given as follows:
\begin{align}
    C_{\mathrm{b},\mathrm{s}} = \mqty( C_{\mathrm{b},\mathrm{m}} & C_{\mathrm{b},\mathrm{ct}} ) =
    \left(\scalebox{0.93}{\begin{tabular}{@{}r@{\,}|@{\,}r@{\,}r@{\,}r@{\,}r@{\,}r@{\,}|@{\,}r@{\,}r@{\,}r@{\,}r@{\,}r@{\,}r@{\,}r@{\,}r@{\,}r@{\,}r@{\,}r@{\,}r@{\,}r@{}}
        & $\ell_1$ & $\ell_2$ & $\ell_3$ & $\ell_4$ & $\ell_5$ & $\ell_6$ & $\ell_7$ & $\ell_8$ & $\ell_9$ & $\ell_{10}$ & $\ell_{11}$ & $\ell_{12}$ & $\ell_{13}$ & $\ell_{14}$ & $\ell_{15}$ & $\ell_{16}$ & $\ell_{17}$ & $\ell_{18}$ \\\hline
        B0 & 1 & 0 & 0 & 0 & 0 & 0 & 0 & 0 & 0 & 0 & 0 & 0 & 0 & 0 & 0 & 0 & 0 & 0  \\
        B1 & 0 & 1 & 0 & 0 & 0 & 1 & 0 & 0 & 0 & 0 & 0 & 0 & 0 & 0 & 0 & 0 & 0 & 0  \\
        B2 & 0 & 1 & 0 & 0 & 0 & 0 & 1 & 0 & 0 & 0 & 0 & 0 & 0 & 0 & 0 & 0 & 0 & 0  \\
        B3 & 0 & 0 & 1 & 0 & 0 & 0 & 0 & 1 & 0 & 0 & 0 & 0 & 0 & 0 & 1 & 0 & 0 & 1  \\
        B4 & 0 & 0 & 1 & 0 & 0 & 0 & 0 & 0 & 1 & 0 & 0 & 0 & 0 & 0 & 0 & 1 & 0 & 0  \\
        B5 & 0 & 0 & 0 & 1 & 0 & 0 & 0 & 0 & 0 & 1 & 0 & 0 & 0 & 0 & 1 & 0 & 0 & 1  \\
        B6 & 0 & 0 & 0 & 1 & 0 & 0 & 0 & 0 & 0 & 0 & 1 & 0 & 0 & 0 & 0 & 1 & 0 & 0  \\
        B7 & 0 & 0 & 0 & 1 & 0 & 0 & 0 & 0 & 0 & 0 & 0 & 1 & 0 & 0 & 0 & 0 & 1 & 1  \\
        B8 & 0 & 0 & 0 & 0 & 1 & 0 & 0 & 0 & 0 & 0 & 0 & 0 & 1 & 0 & 0 & 1 & 0 & 0  \\
        B9 & 0 & 0 & 0 & 0 & 1 & 0 & 0 & 0 & 0 & 0 & 0 & 0 & 0 & 1 & 0 & 0 & 1 & 1  \\
    \end{tabular}}\right)
    \in \mathbb{R}^{{n_\mathrm{j}}\times{n_\mathrm{s}}}.
    \label{TMP_YVOYondqiyn8}
\end{align}

\begin{remark}
Although the length of segments are computed for junctions fixed to the bones, this modeling error is vanishingly small in ${L_\mathrm{b}({\color{black} \theta})}$ as the length of segments are accumulated along a tendon branch.
\end{remark}

\subsection{Constants and measured variables}
First of all, the tensions ${f_\mathrm{m}} \in \mathbb{R}^{n_\mathrm{m}}$ acting in tendons directly connected to the muscles can be obtained from the servo motor armature current, or by using custom tension sensors, e.g., \cite{2017_Niehues}.
Secondly, the lengths of tendon segments ${\Delta L_\mathrm{m}} \in \mathbb{R}^{n_\mathrm{m}}$ coiled on the motor shaft can be inferred from the servo motor's encoder.
Whereas, Young's modulus ${E}$, the cross section area ${A}$, and hence the spring coefficients ${k_i} = \frac{{E} \cdot {A}}{{L_{\mathrm{s}0,i}}}$ of the tendon segments can be measured apriori, $i \in \qty{1,\dots,n_\mathrm{s}}$.

\subsection[Unknown variables]{{Unknown} variables}
Since no encoders are mounted in the finger joints, the angles ${\theta} \in \mathbb{R}^{n_\theta}$ of the free joints are all unknown.
Moreover, the actual excursion ${\ell_\mathrm{ct}} \in \mathbb{R}^{n_\mathrm{ct}}$, and hence the acting tensions ${f_\mathrm{ct}} = {K_\mathrm{ct}} {\ell_\mathrm{ct}}$ in the connected tendons are unknown.
Diagonal matrix ${K_\mathrm{ct}}$ contains the spring coefficients of the connected tendon segments, i.e., ${K_\mathrm{ct}} = \mathrm{diag}({k_{n_\mathrm{m}+1}},\dots,{k_{n_\mathrm{m}+n_\mathrm{ct}}})$.

The measured an unknown variables are all illustrated in Figures \ref{fig:pest_problem} and \ref{fig:branches}.

\subsection{Dependent variables}
Our model allows to compute the total excursion of a tendon branch in two different ways. First, we can infer it from the excursions of individual segments, which form the branch, namely:
\begin{equation}
    \label{eq:lb_1}
    \ell_\mathrm{b} = C_{\mathrm{b},\mathrm{s}} \ell_\mathrm{s} = C_{\mathrm{b},\mathrm{m}} {\ell_\mathrm{m}} + C_{\mathrm{b},\mathrm{ct}} {\ell_\mathrm{ct}}.
\end{equation}
Secondly, the fixed-site geometric tendon model makes it possible to compute the total length, and hence the excursion of a branch from the gesture of the hand as follows:
\begin{equation}
    \label{eq:lb_2}
    \underbrace{L_\mathrm{b}({\theta}) + \Delta L_\mathrm{b}}_{\mathclap{\text{stretched length of the branch~~~~~~~~~~~~~~~}}}
    - \overbrace{{L_{\mathrm{b}0}}}^{\mathclap{\text{unstretched length of the branch}}}
    = \underbrace{\ell_\mathrm{b}}_{\mathclap{\text{~~~~~~~~~~~~~~~~~~total excursion along the branch}}}~~~~~~~~~
\end{equation}
where $L_\mathrm{b}({\theta})$ is the length of tendon branch from the shaft to its endpoint where it is attached to the bone, whereas, $\Delta L_\mathrm{b}$ is the length of tendon branch which is ``missing'' compared to its rest position length. This ``missing'' tendon segment of length $\Delta L_\mathrm{b}$ is coiled on the shaft as it is illustrated in Figure \ref{fig:branches}.

Then, an equation from \eqref{eq:lb_1} and \eqref{eq:lb_2} can be obtained as follows:
\begin{align}
    \label{Eq1:Branch_Excursion}
    L_\mathrm{b}({\theta}) \!+\! C_{\mathrm{b},\mathrm{m}} \, {\Delta L_\mathrm{m}} \!-\! {L_{\mathrm{b}0}} \!=\! C_{\mathrm{b},\mathrm{m}} \, \underbrace{K_\mathrm{m}^{-1} \, {f_\mathrm{m}}}_{{\ell_\mathrm{m}}} + C_{\mathrm{b},\mathrm{ct}} \, {\ell_\mathrm{ct}}.
\end{align}
Finally, we exploit the fact that the acting forces in a junction cancel each other out.
E.g., in Figure \ref{fig:tendon_structure_F2}, forces $f_2$, $f_6$, and $f_7$ acting in junction J1 must satisfy $f_2 = f_6 + f_7$. These conditions can be described by the following equation:
\begin{equation}
    \label{Eq2:Junctions}
    C_{\mathrm{j},\mathrm{s}} \, f_\mathrm{s} = C_{\mathrm{j},\mathrm{m}} \, {f_\mathrm{m}} + C_{\mathrm{j},\mathrm{ct}} \, K_\mathrm{ct} {\ell_\mathrm{ct}} = 0,
\end{equation}
where $C_{\mathrm{j},\mathrm{s}}$ represents the connection matrix of tendon junctions. The $(i,j)$th element of $C_{\mathrm{j},\mathrm{s}}$ is $-1$ if segment $j$ enters junction $i$, $+1$ if it leaves, and 0 if they are not connected. At this point we assumed that all forces arising at a junction act along the same line, namely, the angle between every two force vectors are 0 or $\pi$.
The actual value of $C_{\mathrm{j},\mathrm{s}}$ for the long fingers is
\begin{align}
    C_{\mathrm{j},\mathrm{s}} = \mqty( C_{\mathrm{j},\mathrm{m}} & C_{\mathrm{j},\mathrm{ct}} ) =
    \left(\scalebox{0.93}{\begin{tabular}{@{}r@{\,}|@{\,}r@{\,}r@{\,}r@{\,}r@{\,}r@{\,}|@{\,}r@{\,}r@{\,}r@{\,}r@{\,}r@{\,}r@{\,}r@{\,}r@{\,}r@{\,}r@{\,}r@{\,}r@{\,}r@{}}
        & $f_1$ & $f_2$ & $f_3$ & $f_4$ & $f_5$ & $f_6$ & $f_7$ & $f_8$ & $f_9$ & $f_{10}$ & $f_{11}$ & $f_{12}$ & $f_{13}$ & $f_{14}$ & $f_{15}$ & $f_{16}$ & $f_{17}$ & $f_{18}$ \\\hline
        J1 & 0 &-1 & 0 & 0 & 0 & 1 & 1 & 0 & 0 & 0 & 0 & 0 & 0 & 0 & 0 & 0 & 0 & 0  \\
        J2 & 0 & 0 &-1 & 0 & 0 & 0 & 0 & 1 & 1 & 0 & 0 & 0 & 0 & 0 & 0 & 0 & 0 & 0  \\
        J3 & 0 & 0 & 0 &-1 & 0 & 0 & 0 & 0 & 0 & 1 & 1 & 1 & 0 & 0 & 0 & 0 & 0 & 0  \\
        J4 & 0 & 0 & 0 & 0 &-1 & 0 & 0 & 0 & 0 & 0 & 0 & 0 & 1 & 1 & 0 & 0 & 0 & 0  \\
        J5 & 0 & 0 & 0 & 0 & 0 & 0 & 0 &-1 & 0 &-1 & 0 & 0 & 0 & 0 & 1 & 0 & 0 & 0  \\
        J6 & 0 & 0 & 0 & 0 & 0 & 0 & 0 & 0 &-1 & 0 &-1 & 0 &-1 & 0 & 0 & 1 & 0 & 0  \\
        J7 & 0 & 0 & 0 & 0 & 0 & 0 & 0 & 0 & 0 & 0 & 0 &-1 & 0 &-1 & 0 & 0 & 1 & 0  \\
        J8 & 0 & 0 & 0 & 0 & 0 & 0 & 0 & 0 & 0 & 0 & 0 & 0 & 0 & 0 &-1 & 0 &-1 & 1  \\
    \end{tabular}}\right)
    \in \mathbb{R}^{{n_\mathrm{j}}\times{n_\mathrm{s}}}.
    \label{TMP_BOalhVXHbGWg}
\end{align}


\subsection{The nonlinear program}

The number of equations in \eqref{Eq1:Branch_Excursion} and \eqref{Eq2:Junctions} are $n_\mathrm{b}$ and $n_\mathrm{j}$, respectively. Whereas, the number of unknown variables is $n_\theta + n_\mathrm{ct}$. Moreover, it is reasonable to assume that the number of muscle actuators ($n_\mathrm{m}$) of a finger is at least equal to the number of free joints $n_\theta$, i.e.,
\begin{equation}
    \label{eq:ntheta_le_nm}
    n_\theta \le n_\mathrm{m}.
\end{equation}
If we add $n_\mathrm{ct} = n_\mathrm{s} - n_\mathrm{m} = n_\mathrm{b} + n_\mathrm{j} - n_\mathrm{m}$ to both sides of inequality \eqref{eq:ntheta_le_nm}, we obtain
\begin{equation}
    \label{eq:ntheta_nct_le_nb_nj}
    n_\theta + n_\mathrm{ct} \le n_\mathrm{b} + n_\mathrm{j}.
\end{equation}
Finally, we can conclude that the number of equations in \eqref{Eq1:Branch_Excursion} and \eqref{Eq2:Junctions} is at least the number of unknown with the assumption that $n_\theta \le n_\mathrm{m}$.

The nonlinear program to estimate the gesture $\theta$ can summarized as follows:
\begin{problem}
Assume that the lengths $\Delta L_\mathrm{m}$ of tendon segments coiled on the motors' shafts are available as well as the tensile forces $f_\mathrm{m}$ are measurable at the shafts. Assume also that the length of tendon branches can be explicitly expressed by the joint angles through the differentiable function $L_\mathrm{b}$, hence, the length of branches in the rest position ($L_{\mathrm{b}0}$) are also known.
We are looking for the joint angles $\theta$ and string elongations $\ell_\mathrm{ct}$ in the connected (hidden) tendon segments, which satisfy the identities \eqref{Eq1:Branch_Excursion} and \eqref{Eq2:Junctions}:
\begin{align*}
    \text{Eq. } \eqref{Eq1:Branch_Excursion}: &&
    L_\mathrm{b}({\theta}) + C_{\mathrm{b},\mathrm{m}} \, {\Delta L_\mathrm{m}} - {L_{\mathrm{b}0}} &= C_{\mathrm{b},\mathrm{m}} \, K_\mathrm{m}^{-1} \, {f_\mathrm{m}} + C_{\mathrm{b},\mathrm{ct}} \, {\ell_\mathrm{ct}}.
\\
    \text{Eq. } \eqref{Eq2:Junctions}: &&
    C_{\mathrm{j},\mathrm{s}} \, f_\mathrm{s} &= C_{\mathrm{j},\mathrm{m}} \, {f_\mathrm{m}} + C_{\mathrm{j},\mathrm{ct}} \, K_\mathrm{ct} {\ell_\mathrm{ct}} = 0,
\end{align*}
where matrices $C_{\mathrm{b},\mathrm{m}}$, $C_{\mathrm{b},\mathrm{ct}}$, $C_{\mathrm{j},\mathrm{s}}$, $C_{\mathrm{j},\mathrm{m}}$, $C_{\mathrm{j},\mathrm{ct}}$, and the diagonal $K_\mathrm{m}$ are all known and constants.
\end{problem}

To solve \eqref{Eq1:Branch_Excursion} and \eqref{Eq2:Junctions}, we executed a constrained gradient-based search without a cost function.
For algorithmic differentiation, we used CasADi \cite{2018_Andersson.etal}. To solve the non-linear feasibility problem, we used IPOPT \cite{2005_Waechter.Biegler_IPOPT}, an interior point line search algorithm, with the MUlti-frontal Massively Parallel sparse direct Solver (MUMPS) \cite{2019_Amestoy.etal_MUMPS_Solver,2001_Amestoy.etal_MUMPS_Solver}.

\section{Simulation experiment}
Here, we evaluate a simple posture tracking controller, in which the error between the desired position $\theta^{(\mathrm{d})}$ and the estimated position $\hat\theta$ is fed back through a specific Jacobian matrix.
The experiment is performed in MuJoCo simulation environment \cite{2012_Todorov.etal}.

\subsection{Implementation-specific behavior of the MuJoCo tendon model}
In MuJoCo, tendon junctions are implemented through pulley elements rather than explicit topological branching. Specifically, when a tendon splits into multiple segments using pulley operators, the outgoing segments are interpreted as a single continuous tendon routed through an ideal pulley. As a consequence, the tensile forces acting in all outgoing branches is inherently identical and directly determined by the actuator associated with the root segment. This pulley-based abstraction naturally generalizes to an arbitrary number of connected tendon segments, implying that the tensile forces in all outgoing branches are equal, up to the scaling introduced by the corresponding pulley divisors.
Furthermore, the total tendon length used by the actuator is computed as a weighted sum of the lengths of the individual branches, which, in the present implementation, reduces to the average branch length due to the symmetric pulley configuration.

While MuJoCo provides native support for tendon branching through the use of pulley elements, the re-merging of tendon branches is not straightforward in MuJoCo. That is, once a tendon is split into multiple branches, these branches cannot be recombined into a single path within the same tendon definition. Such kind of tendon connection appears in junctions J5--J8 (Figure \ref{fig:tendon_structure_F2}). Multiple workarounds exist how to mimic general (not tree-like) tendon graphs, e.g., by introducing auxiliary bodies sliding along a bone, which connects multiple tendons. Instead, we decided to model tendon branches independently of each other. Namely, each branch in \eqref{eq:Branches} are terminating independently and the pulley effect of the critical junctions J5--J8 are neglected. However, the branches remain mechanically coupled through the shared actuators (EDC, UI, LUM) and pulley-based force scaling.

Let $L_\mathrm{t} : \mathbb{R}^{n_\theta} \to \mathbb{R}^{n_\mathrm{m}}$ denote the length of the tendons as they are calculcated in the MuJoCo model:
\begin{equation}
    L_\mathrm{t}(\theta) = C_{\mathrm{t},\mathrm{b}} \, L_\mathrm{b}(\theta) = C_{\mathrm{t},\mathrm{b}} \, C_{\mathrm{b},\mathrm{s}} \, L_\mathrm{s}(\theta),
\end{equation}
where matrix
\begin{align}
    \label{TMP_4p5jDvJJNN40}
    &C_{\mathrm{t},\mathrm{b}} = \left(\scalebox{0.93}{\begin{tabular}{@{}c|ccccccccccc@{}}
        & B0 & B1 & B2 & B3 & B4 & B5 & B6 & B7 & B8 & B9 \\\hline
        FDP & 1 & 0   & 0   & 0   & 0   & 0   & 0   & 0   & 0   & 0   \\
        FDS & 0 & 1/2 & 1/2 & 0   & 0   & 0   & 0   & 0   & 0   & 0   \\
        UI  & 0 & 0   & 0   & 1/2 & 1/2 & 0   & 0   & 0   & 0   & 0   \\
        EDC & 0 & 0   & 0   & 0   & 0   & 1/3 & 1/3 & 1/3 & 0   & 0   \\
        LUM & 0 & 0   & 0   & 0   & 0   & 0   & 0   & 0   & 1/2 & 1/2 \\
    \end{tabular}}\right) \in \mathbb{R}^{{n_\mathrm{m}}\times{n_\mathrm{b}}}.
\end{align}
implements the mean length of branches of the tendons.

\subsection{Feedback design for gesture realization through tendon excursion}
In this section, we derive a feedback control law that enables to execute target gestures through appropriate modulation of tendon excursion. The relationship between infinitesimal variations of the joint configuration and the corresponding changes in tendon length is described by the moment arm matrix, defined as
\begin{equation}
    \label{eq:Jacobian}
    R(\theta) = \pdv{L_\mathrm{t}}{\theta}(\theta) \in \mathbb{R}^{n_\mathrm{m} \times n_\theta}.
\end{equation}
Accordingly, differential changes in tendon length satisfy
\begin{equation}
    \label{TMP_a31SbOim0Kal}
    \diffd L_\mathrm{t}(\theta) = \pdv{L_\mathrm{t}}{\theta}(\theta)\, \diffd \theta
    = R(\theta)\, \diffd \theta.
\end{equation}

The modulation of tendon excursion can be realized by coiling the root segments on the motor shaft. The lengths of segments coiled up was denoted by $\Delta L_\mathrm{m}$. Similarly, let $\Delta L_\mathrm{t}^{+}$ denote the excursion level in the subsequent time step. If the length coiled on the shaft is increasing, the tendon's length is decreasing (i.e., $\diffd L_\mathrm{t}(\theta) \approx -(\Delta L_\mathrm{t}^{+} - \Delta L_\mathrm{t})$.
Therefore, according to the differential rule \eqref{TMP_a31SbOim0Kal}, the change in the gesture in a time step (from $\theta$ to $\theta^+$) can be estimated as follows:
\begin{equation}
    \Delta L_\mathrm{t}^{+}
    - \Delta L_\mathrm{t}
    \approx
    - R(\theta) (\theta^+ - \theta).
\end{equation}
Let the gesture $\theta^+$ in the next time step be the target state $\theta^{(\mathrm{d})}$. Also remember that the current state is not measured, therefore, instead of $\theta$ we should use its estimate $\hat \theta$. These will leads us to a simple feedback rule:
\begin{equation}
    \Delta L_\mathrm{t}^{+}
    :=
    \Delta L_\mathrm{t}
    - R(\theta) \, \tilde{\theta}.
\end{equation}
where $\tilde{\theta} = \theta^{(\mathrm{d})} - \hat{\theta}$ denotes the error between the targeted and the estimated (observed) gesture.

We may also feed back the cumulative error hence obtaining a proportional--integral (PI) controller, with specific proportional and integral gains $K_\textsc{p}$ and $K_\textsc{i}$ as described below:
\begin{equation}
    \label{eq:PI_tendon_control}
    \Delta L_\mathrm{t}^{+}
    =
    \Delta L_\mathrm{t}
    -
    \underbrace{
        R(\theta)
        \left(
            K_\textsc{p}\, \tilde{\theta}
            +
            K_\textsc{i}\, \sum_{k} h\, \tilde{\theta}(k)
        \right)
    }_{\text{PI feedback}},
\end{equation}
where $h$ is the sampling period used in the error accumulation (integration).
%

In the simulations, the integral gain $K_\textsc{i}$ was selected to be approximately one order of magnitude smaller than the proportional gain
$K_\textsc{p}$ in order to ensure stable convergence while avoiding excessive integral windup.

\subsection{Feedforward term}
Assume that in each time step, we can estimate the necessary tendon excursion, which can lead the hand to the desired gesture at that moment. Let us denote this estimated actuator length by $\Delta \widehat L_\mathrm{t}$. Additionally, we use the notation $\Delta \widehat L_\mathrm{t}^-$ to refer to the estimated actuator length in the previous step. Then, the control can be supplemented with a feedforward term as follows:
\begin{equation}
    \label{eq:ffwd_fbck}
    \Delta L_\mathrm{t}^+ = \Delta L_\mathrm{t}
    +
    \underbrace{
        \pqty\big{\Delta \widehat L_\mathrm{t} - \Delta \widehat L_\mathrm{t}^-}
    }_\text{feedforward term}
    -
    \underbrace{
        R(\theta) \, \qty( K_\textsc{p} \, \tilde \theta + K_\textsc{i} \, {\textstyle \sum_{k}} h \, \tilde \theta(k) ),
    }_\text{PI feedback}
\end{equation}
In the ideal case, the feedforward term alone can achieve the control goal with zero error.
Assume that the required tendon excursion at time $k$ is precisely $\Delta \widehat L_\mathrm{t}(k)$. Start the hand at rest position, namely, $\Delta \widehat L_\mathrm{t}(0) = 0$ and set $K_p = K_I = 0$, then we obtain the following control sequence:
\begin{align}
    \begin{aligned}
    &
    \Delta L_\mathrm{t}(0) = 0,
    \\&
    \Delta L_\mathrm{t}(1) = \Delta L_\mathrm{t}(0)
        + \qty\big(\Delta \widehat L_\mathrm{t}(1) - \Delta \widehat L_\mathrm{t}(0))
        = \Delta \widehat L_\mathrm{t}(1),
    \\&
    \Delta L_\mathrm{t}(2) = \Delta L_\mathrm{t}(1)
        + \qty\big(\Delta \widehat L_\mathrm{t}(2) - \Delta \widehat L_\mathrm{t}(1))
        = \Delta \widehat L_\mathrm{t}(2),
    \\&
    \Delta L_\mathrm{t}(3) = \Delta L_\mathrm{t}(2)
        + \qty\big(\Delta \widehat L_\mathrm{t}(3) - \Delta \widehat L_\mathrm{t}(2))
        = \Delta \widehat L_\mathrm{t}(3), ~~ \dots
    \end{aligned}
\end{align}
which is identical to the presumably required control action $\Delta \widehat L_\mathrm{t}(k)$.

\subsubsection{Prediction of the feedforward control action}
The feedforward term requires an estimate of the tendon excursion $\Delta L_\mathrm{t}$ associated with the desired gesture. One possible approach is to estimate the corresponding tendon lengths directly from the simplified (skeleton-based) kinematic model. In this work, however, a data-assisted strategy is adopted.

Specifically, pre-recorded tendon length configurations are used to interpolate an estimate of the tendon excursion for the current target gesture. The target gestures are selected from a predefined set for which the tendons were manually actuated to realize the corresponding hand postures. For each target gesture, a feasible tendon length realization is therefore available and serves as interpolation data for the feedforward term. This approach provides a simple yet efficient estimate of the required tendon excursion and complements the feedback controller by improving transient response.

\subsection{Discussion and comparative evaluation}

This section provides a detailed evaluation of the proposed observer-based PI control framework augmented with feedforward compensation, with particular emphasis on gesture convergence, tracking accuracy, and the interplay between joint-space and task-space performance metrics.

\begin{figure}
    \centering
    \begin{minipage}{0.32\textwidth}\centering\includegraphics[width=\textwidth,clip,trim={470 50 515 50}]{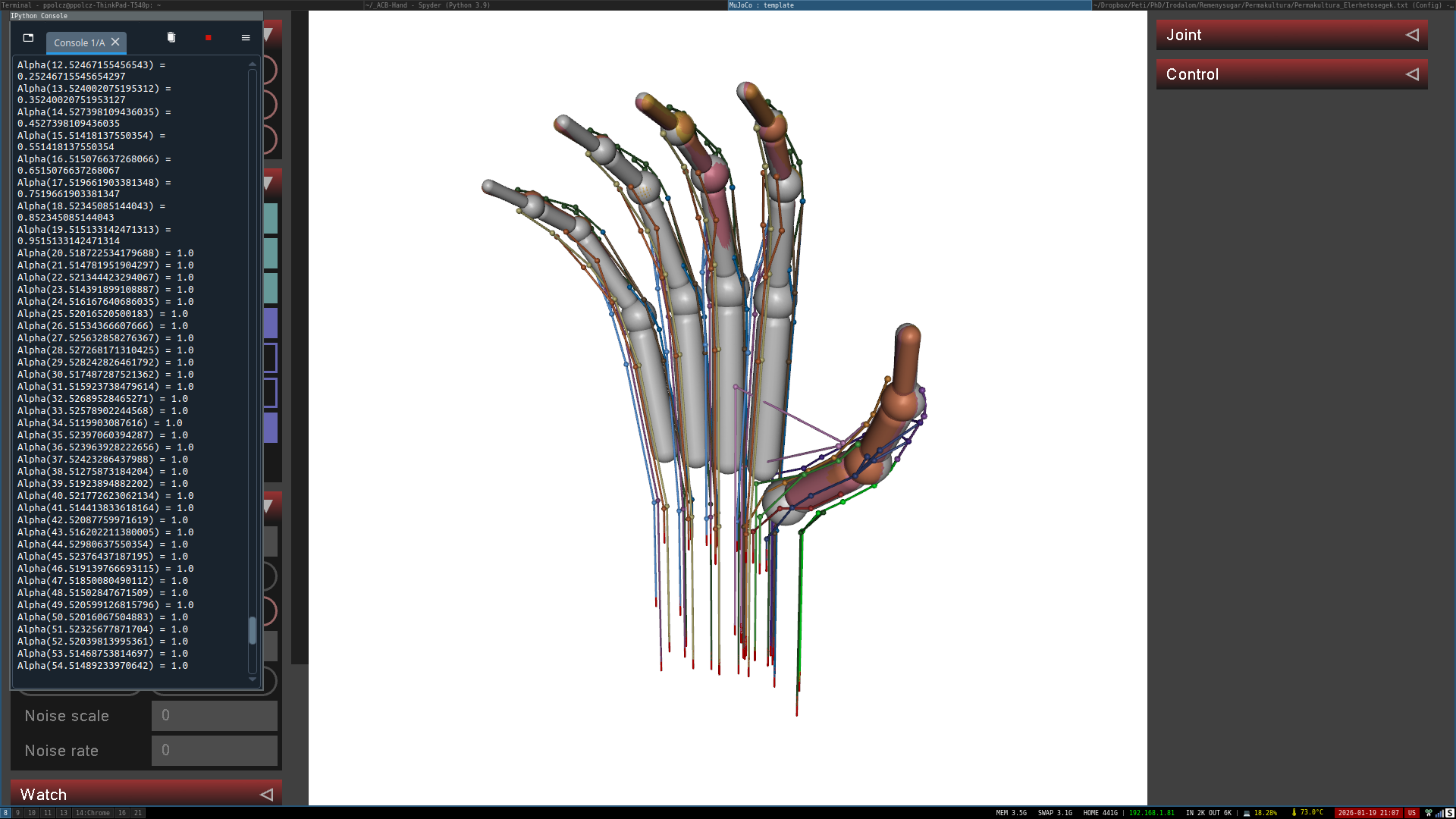} \\ \textbf{Gesture (G1)} \end{minipage}
    \begin{minipage}{0.32\textwidth}\centering\includegraphics[width=\textwidth,clip,trim={470 50 515 50}]{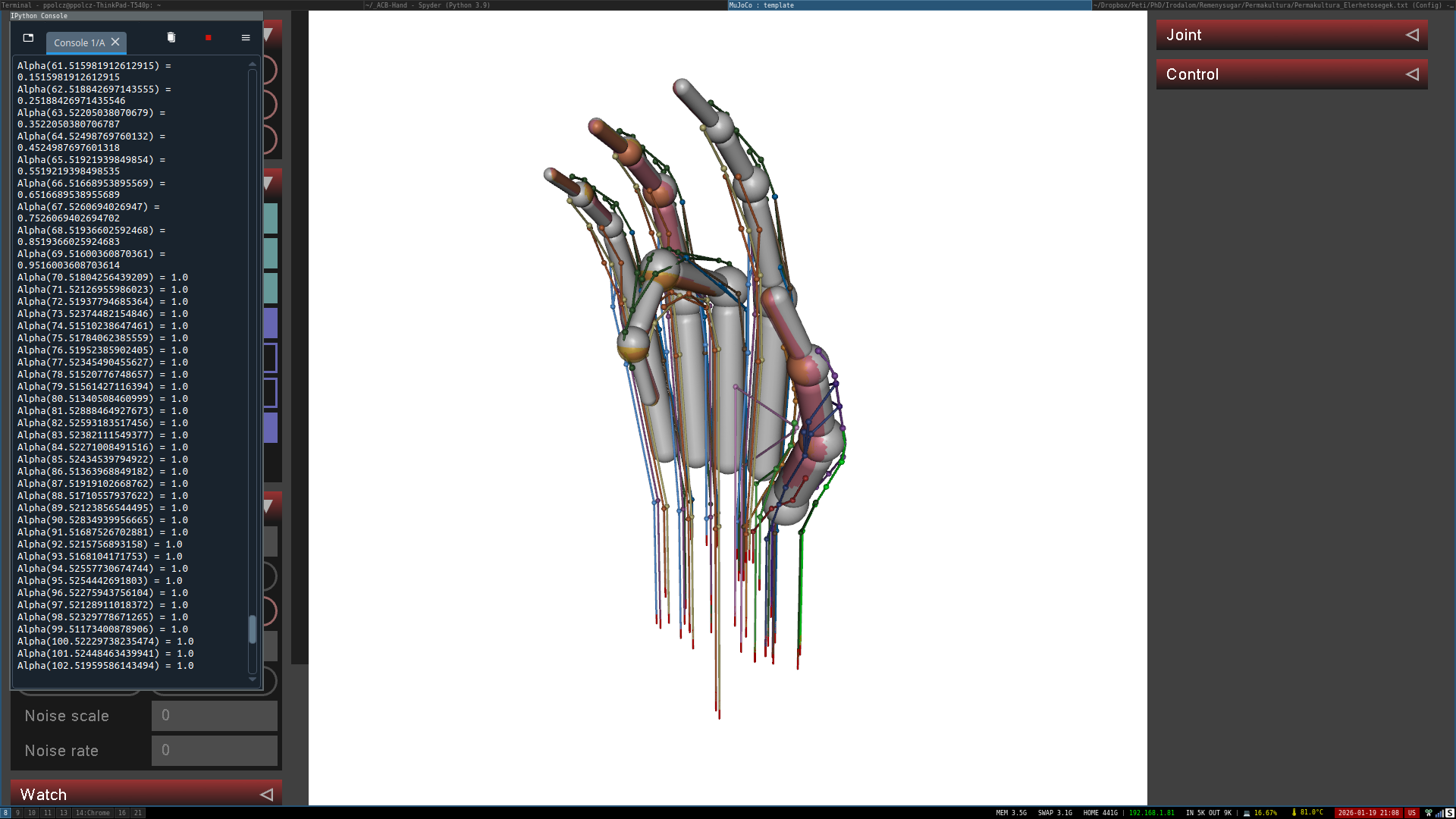} \\ \textbf{Gesture (G2)} \end{minipage}
    \begin{minipage}{0.32\textwidth}\centering\includegraphics[width=\textwidth,clip,trim={470 50 515 50}]{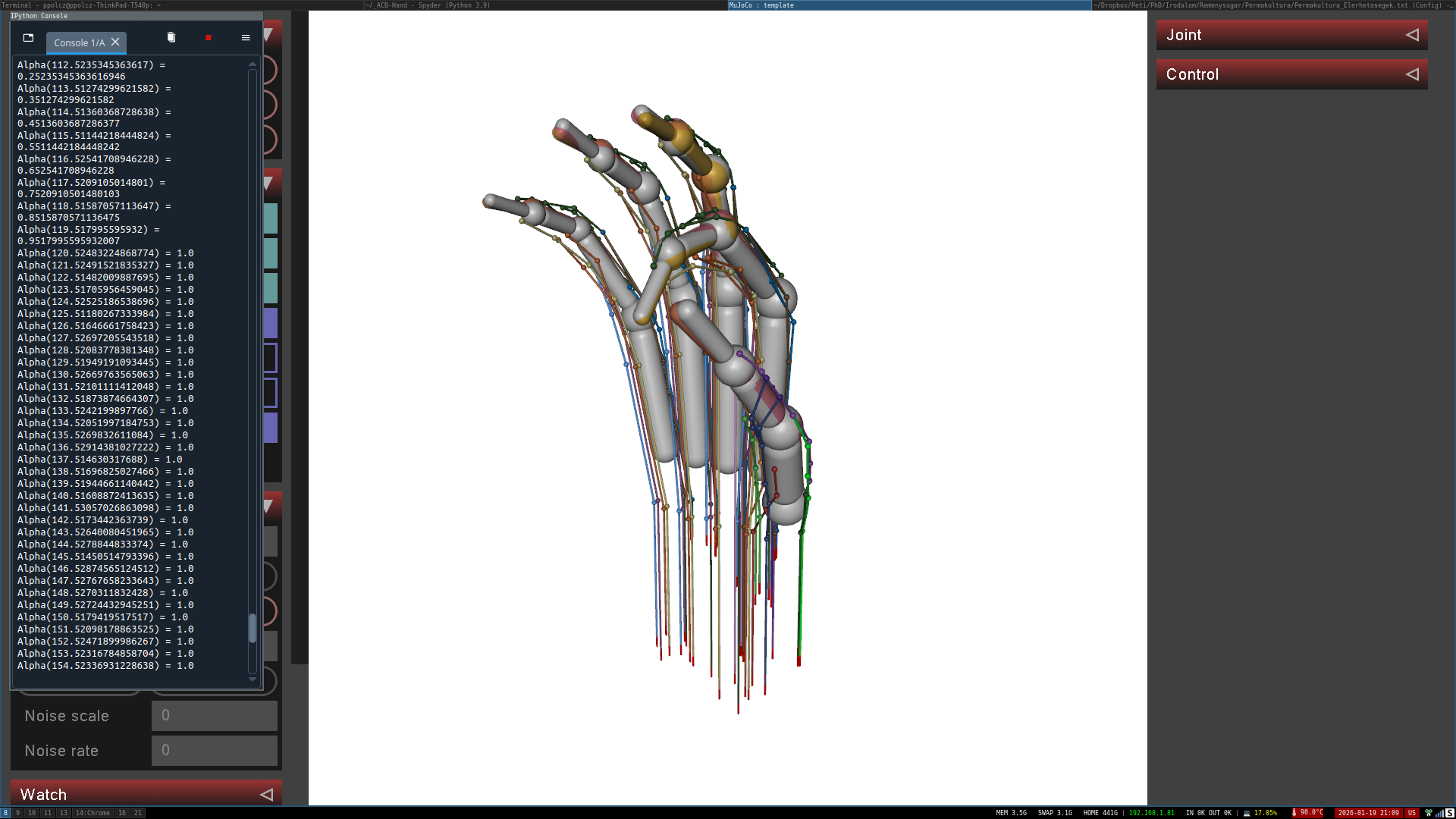} \\ \textbf{Gesture (G3)} \end{minipage}
    \begin{minipage}{0.32\textwidth}\centering\includegraphics[width=\textwidth,clip,trim={470 50 515 50}]{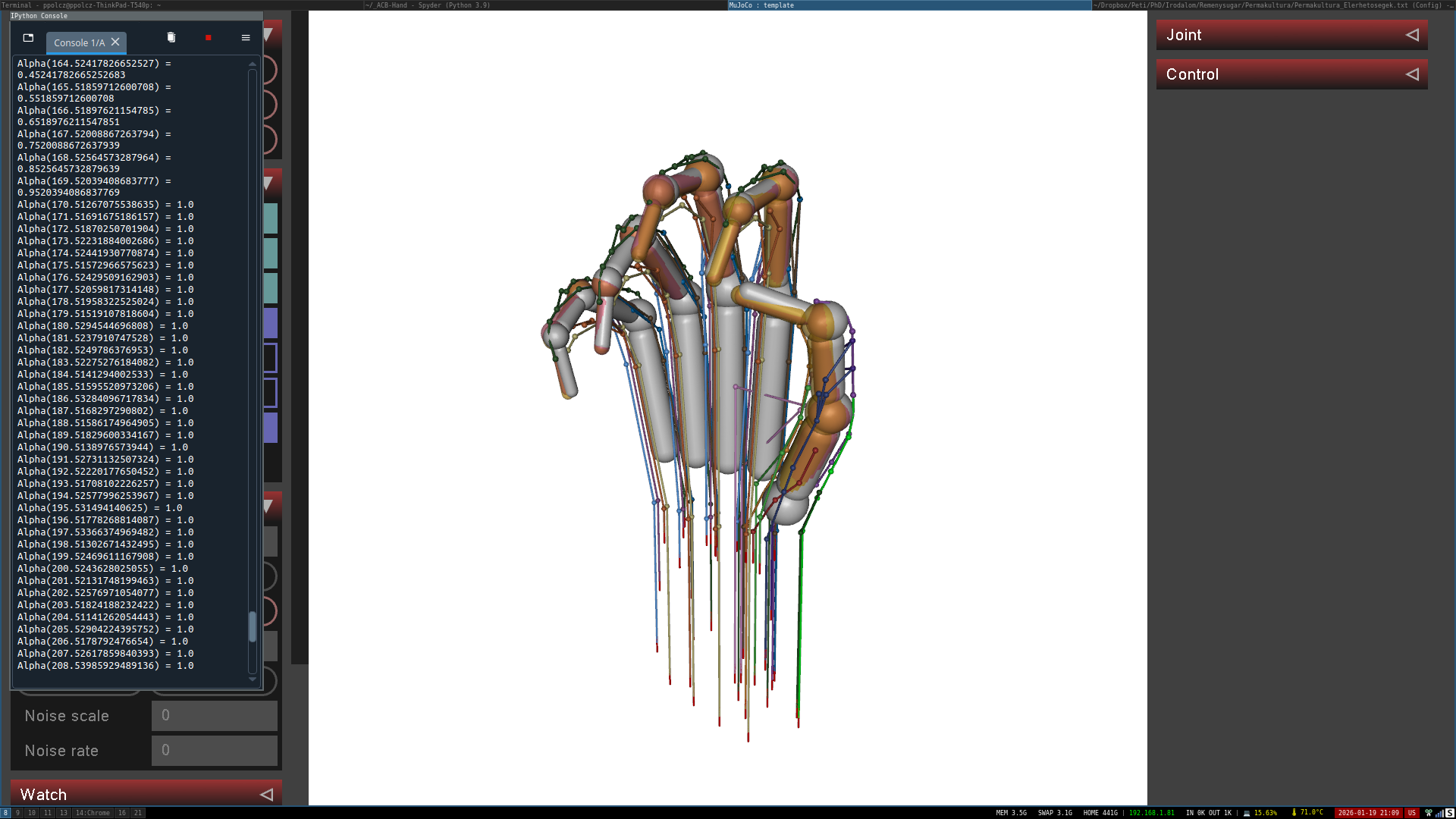} \\ \textbf{Gesture (G4)} \end{minipage}
    \begin{minipage}{0.32\textwidth}\centering\includegraphics[width=\textwidth,clip,trim={470 50 515 50}]{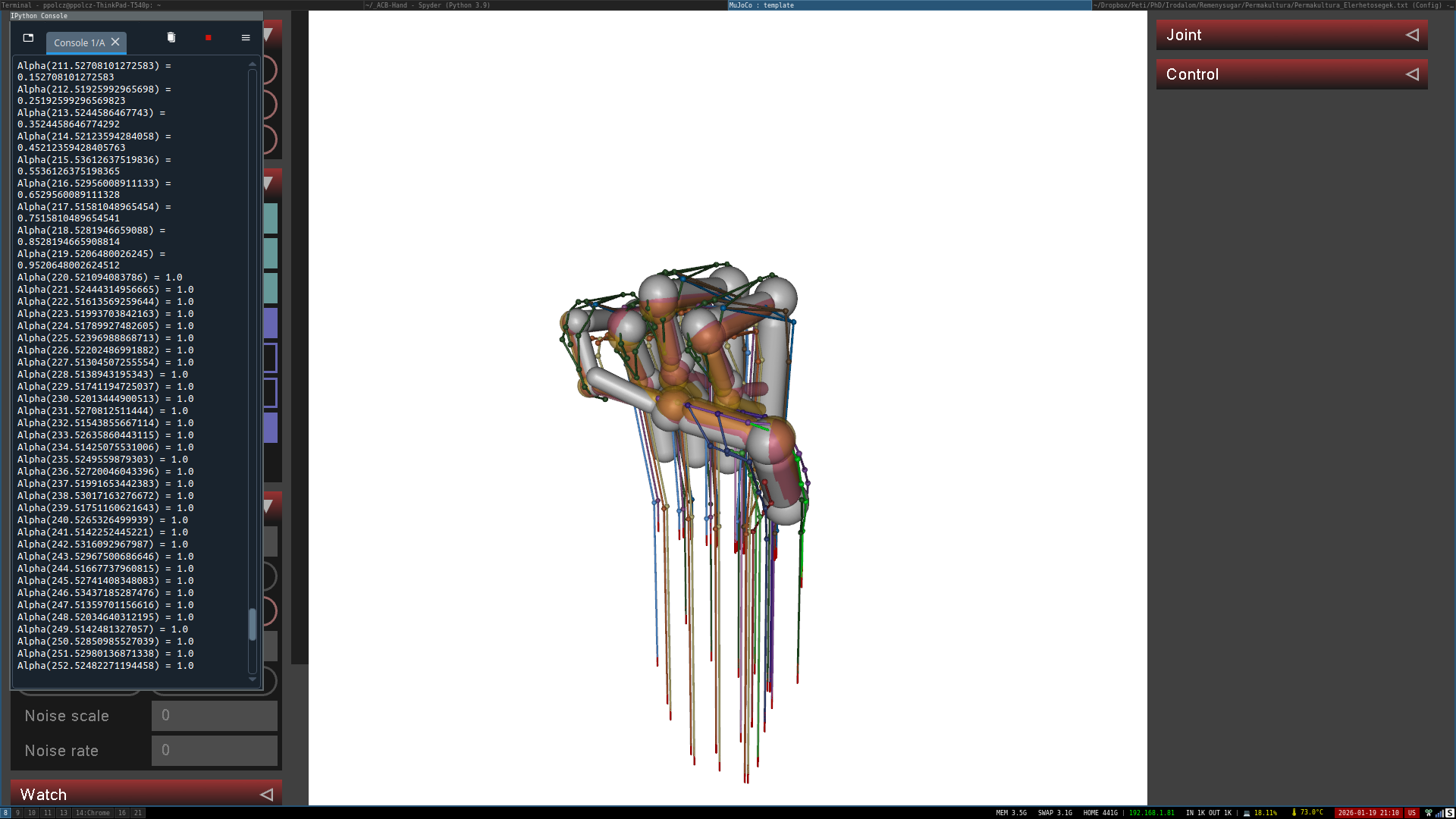} \\ \textbf{Gesture (G5)} \end{minipage}
    \begin{minipage}{0.32\textwidth}\centering\includegraphics[width=\textwidth,clip,trim={470 50 515 50}]{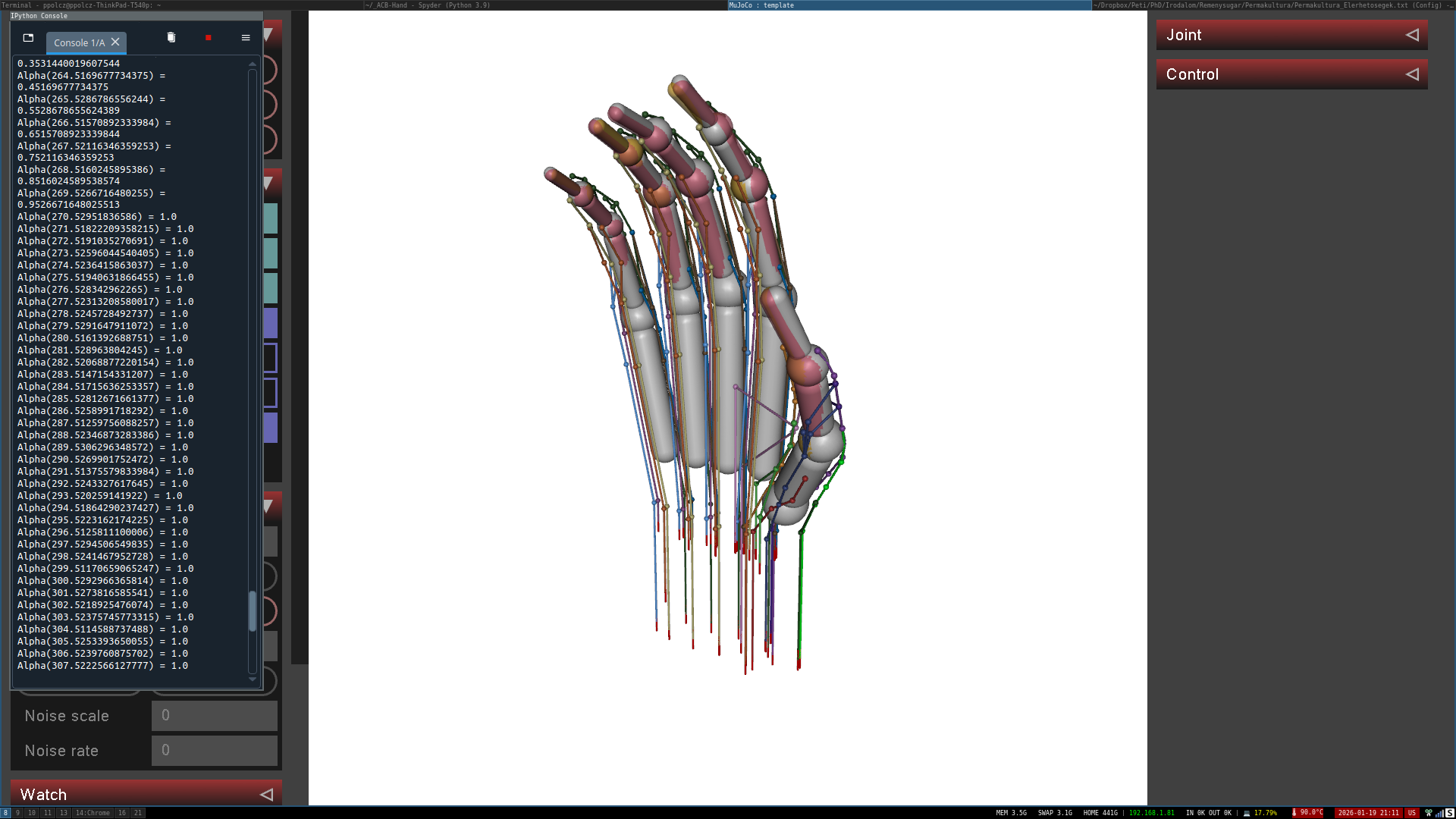} \\ \textbf{Gesture (G6)} \end{minipage}

    \caption{Desired, achieved, and observed gestures under observer-based PI control with feedforward compensation. Red phantom bodies represent the desired gestures, grey bodies correspond to the achieved gestures produced by observer-based PI feedback with an additional feedforward term, and yellow phantom bodies depict the observed gestures. With the exception of Gesture G5, the desired, achieved, and observed gestures exhibit only minor discrepancies, making them nearly indistinguishable to the human eye.}
    \label{fig:Result}
\end{figure}

The overall objective of the presented experiments was to assess whether targeted hand gestures can be reliably achieved using tendon excursion control in the presence of strong kinematic coupling, estimation uncertainty (on the controller's side), and soft joint constraints (on the simulator's side). To this end, six representative gestures were executed sequentially, starting from a resting posture, transitioning smoothly between gestures, and finally returning to rest. The results are evaluated qualitatively through visual comparison (Figure \ref{fig:Result}) and quantitatively through joint-space and task-space error metrics, settling-time analysis, and constraint-violation statistics.

\subsubsection{Joint-space tracking and observability characteristics}

\begin{figure*}
    \centering
    \includegraphics[width=\textwidth]{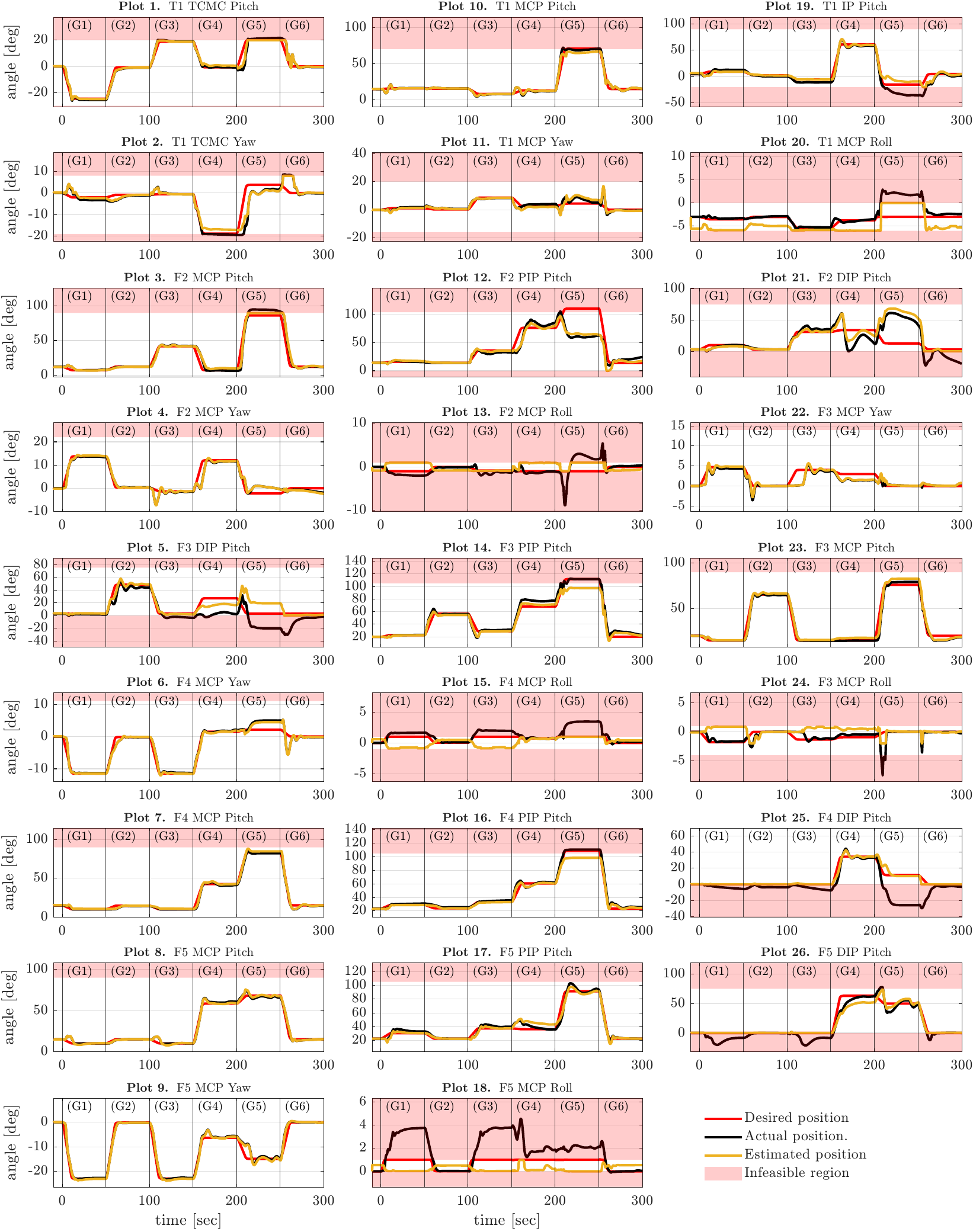}
    \caption{Posture estimation during the PI-based tracking controller with feedforward term. Red lines show to the desired angles, the black illustrate achieved angles, the yellow highlight the estimated angles. The labels (G1)--(G6) in the top of each plot highlight the current gesture to be achieved.}
    \label{fig:Results_XdXoX}
\end{figure*}

\begin{figure*}
    \centering
    \includegraphics[width=\textwidth]{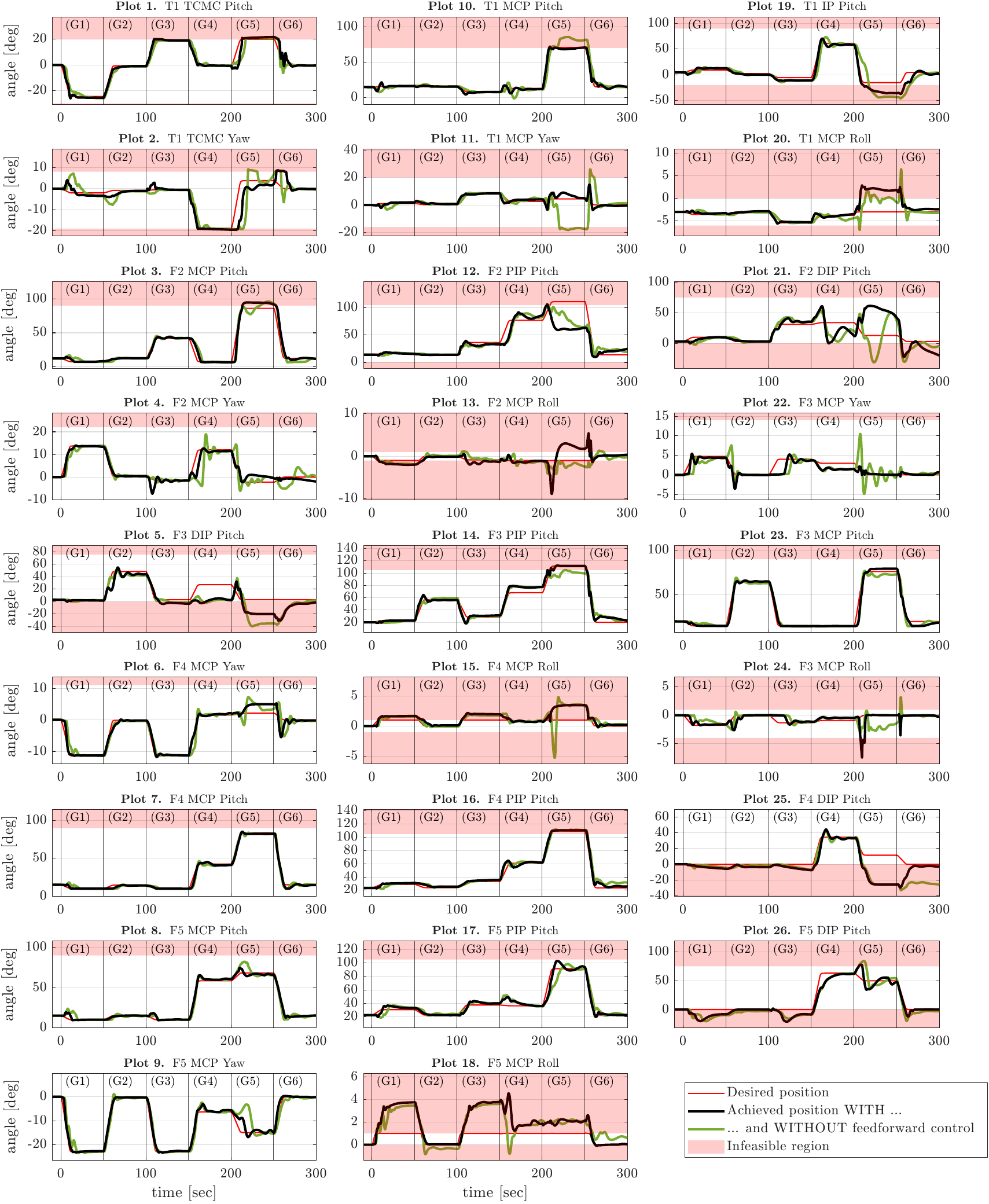}
    \caption{Posture tracking using the PI-based controller with and without feedforward compensation. Red lines show to the desired angles, the black illustrate achieved angles with compensation, the green lines show the achieved angles without compensation.}
    \label{fig:Results_XdX_with_without}
\end{figure*}

The results in Figures~\ref{fig:Results_XdXoX} and~\ref{fig:Results_XdX_with_without} indicate that the proposed observer-based PI controller is capable of tracking most joint angles with good accuracy across a wide range of gestures. For the majority of joints, both the estimated and achieved angles remain close to the desired trajectories, suggesting favorable observability and controllability properties of the tendon-driven system.

Nevertheless, systematic discrepancies are observed in certain rotational degrees of freedom, most notably in the Roll angles and in distal pitch joints (IP and DIP). These joints exhibit larger estimation and tracking errors, which can be attributed to a combination of limited tendon leverage, strong coupling effects, and narrow feasible angle ranges. In particular, the feasible intervals of the MPC Roll angles span only a few degrees, rendering them highly sensitive to modeling inaccuracies and numerical estimation errors. As a result, even small absolute deviations appear pronounced when expressed relative to the admissible range.

Gesture~G5 (clenched fist) represents a particularly challenging configuration. During this gesture, multiple fingers undergo extreme flexion while interacting mechanically with the thumb, leading to intermittent collisions. These contacts, combined with tendon coupling and soft constraint enforcement, introduce local inconsistencies between the estimated and actual joint configurations. Consequently, several joints temporarily enter infeasible regions, as shown in Table~\ref{table:ConViolation}, and joint-angle tracking errors increase noticeably.

\subsubsection{Settling-time analysis in joint space}

\begin{figure*}
    \centering
    \includegraphics[width=0.85\textwidth]{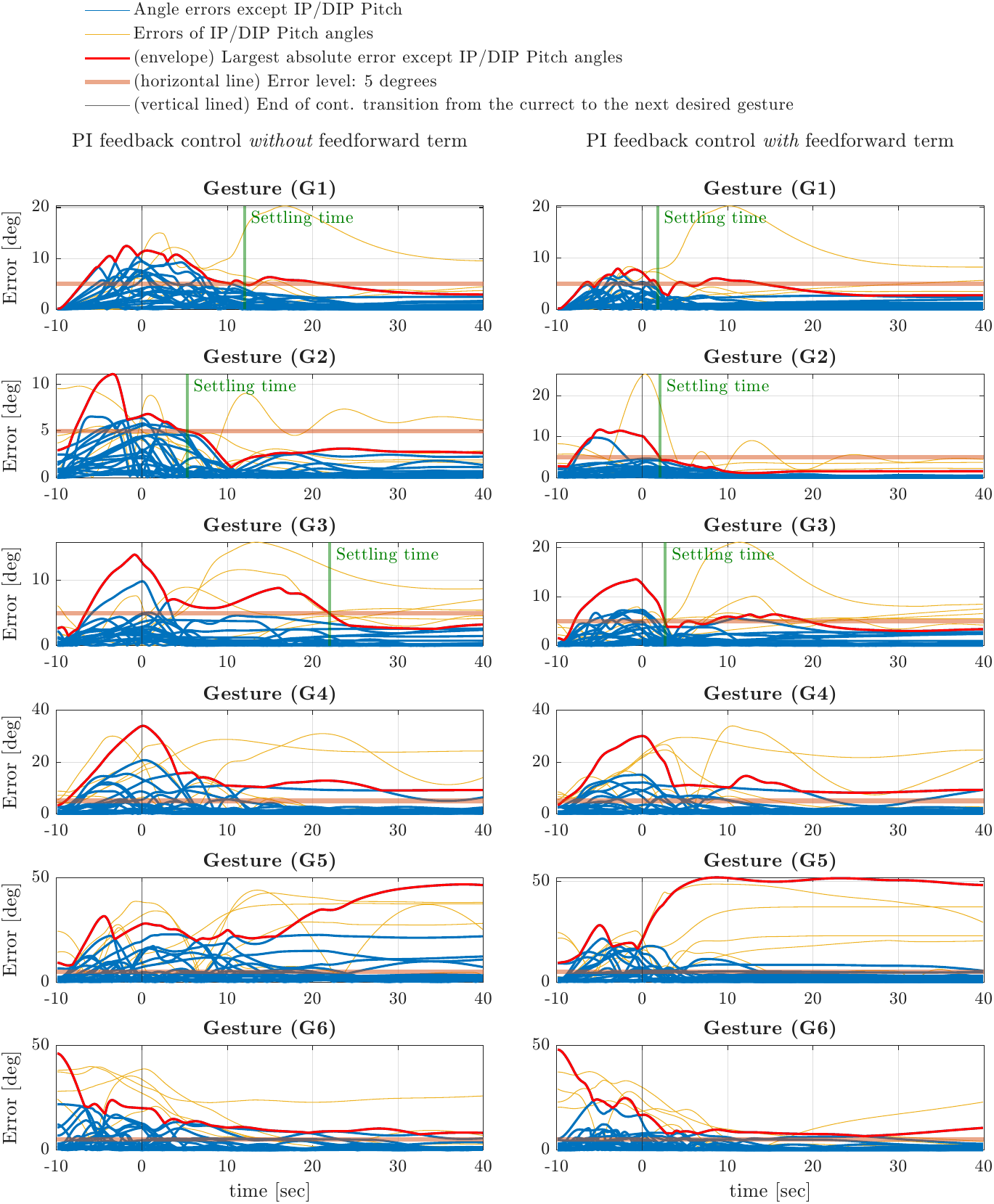}
    \caption{Time evolution of joint-angle tracking errors and settling-time determination for gesture transitions.
For each target gesture, the figure shows the absolute tracking errors of all 26 joint angles with respect to the desired configuration. Errors of all joints except the IP and DIP pitch angles are shown as blue curves, while errors corresponding to IP and DIP pitch angles are shown in yellow. The red curve denotes the envelope of the largest absolute joint error among all joints except the IP/DIP pitch angles. The thick, semi-transparent red horizontal line indicates the error threshold of $5^\circ$ used to define gesture convergence. The black vertical line at $t = 0$ marks the end of the continuous reference transition between consecutive gestures, while the green vertical line indicates the settling time, defined as the first time instant after completion of the reference transition at which the maximum joint-angle error falls below $5^\circ$. The left column shows results obtained with PI feedback only, whereas the right column includes additional feedforward compensation. Due to the constraint violation of the DIP Pitch angles of the long fingers and IP Pitch angle of the thumb, the deviation of these angles are considered when determining the settling time.}
    \label{fig:Results_Settling_Angle}
\end{figure*}

Figure~\ref{fig:Results_Settling_Angle} provides a detailed settling-time analysis based on joint-angle errors. By monitoring the envelope of the largest absolute joint error (excluding IP/DIP pitch angles), gesture convergence is defined in a principled and conservative manner. The use of a linearly interpolated reference trajectory is essential in this context, as it prevents excessive overshoot induced by the proportional feedback term and yields smoother transients.

The comparison between proportional-integral feedback with and without feedforward compensation reveals a clear improvement in transient performance when feedforward is included. In particular, the settling time is significantly reduced for the first three gestures, where the error envelope drops below the $5^\circ$ threshold shortly after completion of the reference transition. However, for the remaining gestures -- especially those involving high degrees of flexion -- the joint-space error does not consistently settle below the threshold within the observation window. This highlights the intrinsic difficulty of accurately tracking highly articulated configurations in joint space under tendon-driven actuation.

\subsubsection{Task-space convergence and perceptual relevance}

\begin{figure*}
    \centering
    \includegraphics[width=0.9\textwidth]{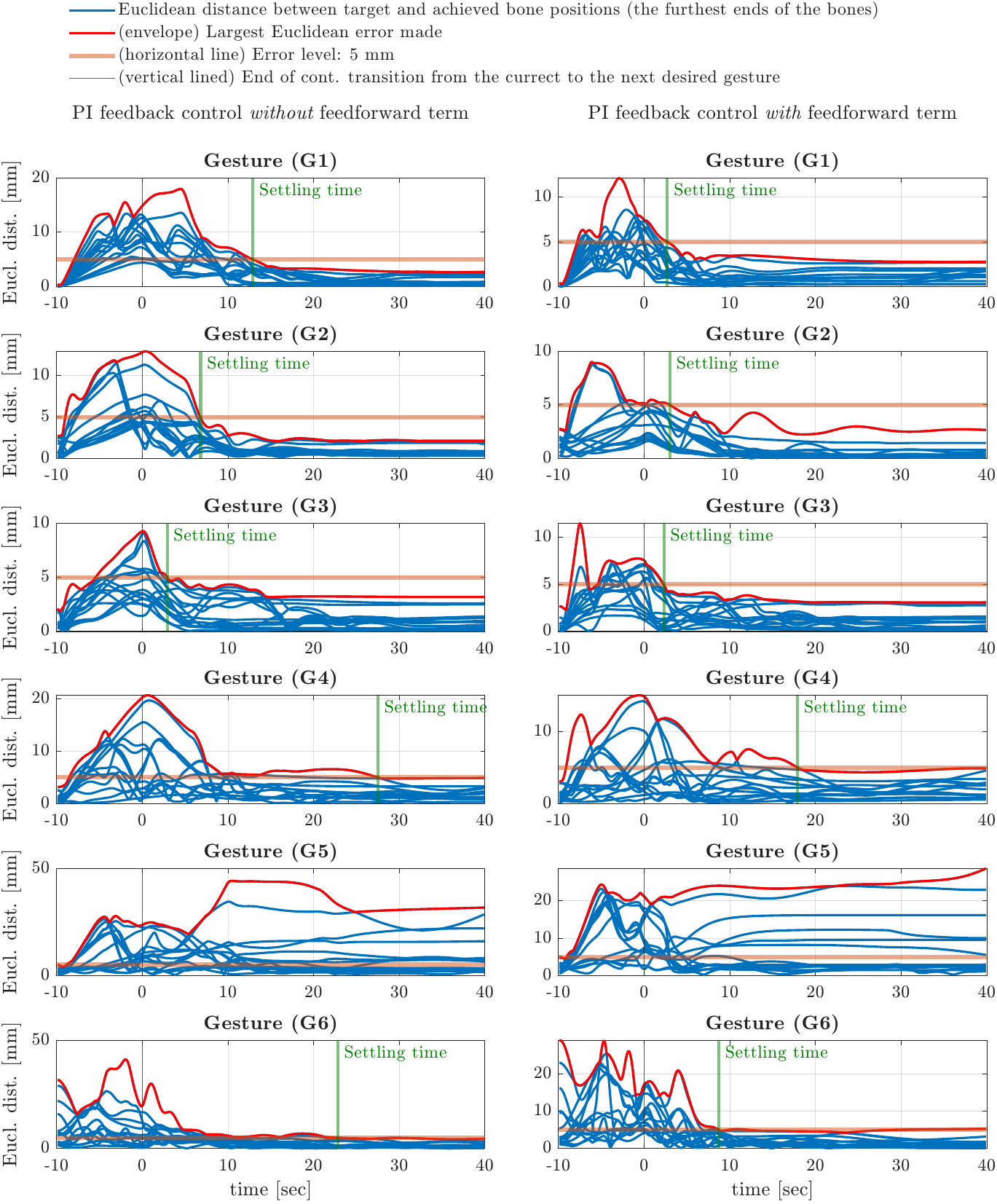}
    \caption{Time evolution of task-space tracking errors and gesture convergence based on fingertip positions.
The figure shows the Euclidean distance between the achieved and desired positions of the distal endpoints of each finger bone during transitions between consecutive gestures. Individual task-space errors are shown as blue curves, while the red curve denotes the envelope corresponding to the largest Euclidean error across all monitored bone endpoints. The thick, semi-transparent red horizontal line indicates the task-space error threshold of $5,\mathrm{mm}$ used to define gesture convergence. Black vertical lines at $t = 0$ mark the end of the continuous reference transition between consecutive gestures. In contrast to joint-space analysis, this figure evaluates gesture convergence directly in task space using Cartesian position errors.}
    \label{fig:Results_Settling_WorldPos}
\end{figure*}

A complementary perspective is provided by the task-space analysis shown in Figure~\ref{fig:Results_Settling_WorldPos}, where gesture convergence is evaluated based on Euclidean position errors of fingertip endpoints. In contrast to joint-space metrics, task-space errors fall below the $5,\mathrm{mm}$ threshold for five out of the six gestures. This result is particularly noteworthy, as it demonstrates that acceptable gesture realization can be achieved in Cartesian space even when residual joint-angle errors persist.

This discrepancy between joint-space and task-space performance underscores an important practical insight: for anthropomorphic hands, perceptual and functional gesture quality is often more closely related to fingertip positions than to precise joint-angle matching. As a result, task-space metrics provide a more meaningful assessment of gesture execution quality in many application scenarios, especially those involving human--robot interaction or grasping tasks.

\subsubsection{Steady-state accuracy and role of feedforward compensation}

The steady-state joint tracking errors reported in Table~\ref{table:JointErrors} further support this interpretation. After allowing sufficient time for transient dynamics to decay, most joints exhibit small residual errors, with larger deviations primarily confined to distal joints and gestures requiring extreme tendon excursion. Importantly, although feedforward compensation plays a crucial role in accelerating convergence and improving transient response, its influence on steady-state behavior is negligible. Consequently, the reported steady-state errors predominantly reflect the intrinsic performance of the PI feedback controller.

This observation suggests that feedforward compensation is best viewed as a transient enhancement mechanism rather than as a means of improving ultimate tracking accuracy. Once the system has settled, the closed-loop feedback dynamics dominate the behavior.

\subsubsection{Constraint violations and implications of soft-constraint modeling}

Table~\ref{table:ConViolation} provides a quantitative summary of joint-limit violations occurring during transient gesture execution. These violations arise from the soft-constraint implementation used in the MuJoCo simulation environment, where joint limits are enforced via compliant, penalty-based forces rather than hard constraints. This formulation allows for physically smooth motion but permits temporary excursions beyond the feasible domain.

Most gestures exhibit only minor violations; however, Gesture~G5 again stands out, with significant deviations observed in multiple distal joints, particularly in DIP pitch angles. These violations correlate strongly with increased tracking errors and visual discrepancies, indicating a complex interaction between soft constraint enforcement, tendon coupling, estimation uncertainty, and aggressive transient dynamics. The results highlight the importance of incorporating constraint-aware control and estimation strategies when executing highly articulated gestures in tendon-driven anthropomorphic hands.

\subsubsection{Summary and outlook}

Overall, the presented results demonstrate that the proposed observer-based PI control framework achieves reliable gesture execution across a relatively wide range of configurations, with favorable task-space performance even in the presence of joint-space inaccuracies and soft constraint violations. The analysis reveals clear trade-offs between transient speed, estimation accuracy, and constraint adherence, and emphasizes the value of task-space metrics for evaluating gesture quality.

Future work will focus on improving roll-angle observability, incorporating explicit constraint handling in the control design, and leveraging data-driven or learning-based techniques to enhance both estimation accuracy and feedforward prediction, particularly for complex, collision-prone gestures. Furthermore, we plan to use machine learning approaches to enhance both the posture estimation and the feedforward compensation.


\begin{table}
\centering
\begin{tabular}{|c@{~}c@{~}c|c|c|c|c|c|c|}
\hline
Finger & Joint & Angle & G1 & G2 & G3 & G4 & G5 & G6 \\
\hline
\hline
T1 & TCMC & Pitch & 1.0 & 0.2 & 0.2 & 0.1 & 1.8 & 0.5 \\
T1 & TCMC & Yaw & 1.3 & 0.2 & 0.0 & 0.6 & 1.7 & 0.2 \\
T1 & MCP & Pitch & 0.4 & 0.2 & 0.2 & 1.0 & 0.2 & 0.3 \\
T1 & MCP & Yaw & 0.8 & 0.4 & 0.2 & 1.4 & 0.5 & 0.5 \\
T1 & MCP & Roll & 0.3 & 0.2 & 0.1 & 0.3 & 4.8 & 0.6 \\
T1 & IP & Pitch & 3.5 & 2.2 & \bf{5.6} & 2.7 & \bf{20.5} & 1.9 \\
\hline
\hline
F2 & MCP & Pitch & 1.0 & 0.1 & 0.6 & 0.8 & \bf{5.6} & 0.6 \\
F2 & MCP & Yaw & 0.4 & 0.1 & 0.2 & 0.5 & 1.8 & 1.9 \\
F2 & MCP & Roll & 1.0 & 0.1 & 0.5 & 0.1 & 2.8 & 0.4 \\
F2 & PIP & Pitch & 0.6 & 0.0 & 3.4 & \bf{9.3} & \bf{48.1} & \bf{10.6} \\
F2 & DIP & Pitch & 0.5 & 0.2 & 4.2 & \bf{21.4} & \bf{29.6} & \bf{23.0} \\
\hline
\hline
F3 & MCP & Pitch & 0.1 & 0.3 & 0.4 & 0.3 & 3.1 & 1.5 \\
F3 & MCP & Yaw & 0.3 & 0.0 & 0.3 & 1.5 & 0.0 & 0.6 \\
F3 & MCP & Roll & 0.2 & 0.0 & 0.2 & 0.4 & 0.0 & 0.3 \\
F3 & PIP & Pitch & 0.2 & 0.5 & 2.7 & \bf{9.2} & 0.1 & 3.0 \\
F3 & DIP & Pitch & 1.4 & 4.4 & \bf{6.6} & \bf{24.6} & \bf{23.0} & 4.2 \\
\hline
\hline
F4 & MCP & Pitch & 0.2 & 0.4 & 0.2 & 1.2 & 1.4 & 0.4 \\
F4 & MCP & Yaw & 0.1 & 0.0 & 0.1 & 0.6 & 2.8 & 0.2 \\
F4 & MCP & Roll & 0.7 & 0.0 & 1.0 & 0.0 & 2.5 & 0.2 \\
F4 & PIP & Pitch & 2.0 & 1.6 & 2.4 & 1.1 & 1.3 & 2.3 \\
F4 & DIP & Pitch & \bf{5.7} & 3.7 & \bf{7.5} & 1.3 & \bf{37.3} & 3.0 \\
\hline
\hline
F5 & MCP & Pitch & 0.0 & 0.0 & 0.0 & 2.1 & 2.4 & 0.1 \\
F5 & MCP & Yaw & 0.3 & 0.1 & 0.3 & 0.7 & 0.0 & 0.1 \\
F5 & MCP & Roll & 2.8 & 0.0 & 2.8 & 0.9 & 1.1 & 0.0 \\
F5 & PIP & Pitch & 2.4 & 0.0 & 2.4 & 0.4 & 2.3 & 0.0 \\
F5 & DIP & Pitch & \bf{8.3} & 0.2 & \bf{8.5} & 1.0 & 2.0 & 0.1 \\
\hline
\end{tabular}
\caption{Absolute joint tracking errors evaluated after gesture convergence.
For each finger, joint, and rotational degree of freedom, the table reports the absolute difference (in degrees) between the achieved joint angle and the desired joint configuration, evaluated at a single time instant $0.1\,\mathrm{s}$ before the selection of the next target gesture.
Each gesture was commanded for $50\,\mathrm{s}$, allowing sufficient time for the transient response to decay and for the hand to approach a steady configuration.
Boldface values indicate tracking errors exceeding $5^\circ$. Most of these deviations can be accounted for constraint violation, which is summarized in Table \ref{table:ConViolation}.
Although PI feedback augmented with feedforward compensation was employed in this case study, the feedforward term has a negligible influence on the steady-state behavior; consequently, the reported values primarily reflect the steady-state performance of the feedback controller.}

\label{table:JointErrors}
\end{table}

\begin{table}
\centering
\begin{tabular}{|c@{~}c@{~}c|c|c|c|c|c|c|}
\hline
Finger & Joint & Angle & G1 & G2 & G3 & G4 & \bf{G5} & G6 \\
\hline
\hline
T1 & TCMC & Pitch & -- & -- & 0.2 & -- & 1.8 & 1.8 \\
T1 & TCMC & Yaw & -- & -- & -- & 0.5 & 0.6 & 0.6 \\
T1 & MCP & Pitch & -- & -- & -- & -- & 2.0 & 0.7 \\
T1 & MCP & Yaw & -- & -- & -- & -- & -- & -- \\
T1 & MCP & Roll & -- & -- & -- & -- & 2.9 & 2.6 \\
T1 & IP & Pitch & -- & -- & -- & -- & \bf{15.6} & \bf{16.9} \\
\hline
\hline
F2 & MCP & Pitch & -- & -- & -- & -- & \bf{5.2} & 2.0 \\
F2 & MCP & Yaw & -- & -- & -- & -- & -- & -- \\
F2 & MCP & Roll & 1.0 & 1.0 & 1.0 & 0.6 & \bf{7.9} & 4.4 \\
F2 & PIP & Pitch & -- & -- & -- & -- & 1.3 & -- \\
F2 & \bf{DIP} & Pitch & 0.2 & -- & -- & 0.1 & -- & \bf{22.6} \\
\hline
\hline
F3 & MCP & Pitch & -- & -- & -- & -- & -- & -- \\
F3 & MCP & Yaw & -- & -- & -- & -- & -- & -- \\
F3 & MCP & Roll & -- & -- & -- & -- & 3.5 & -- \\
F3 & PIP & Pitch & -- & -- & -- & -- & \bf{7.3} & \bf{6.6} \\
F3 & \bf{DIP} & Pitch & 0.3 & -- & 3.7 & 3.6 & \bf{20.0} & \bf{30.5} \\
\hline
\hline
F4 & MCP & Pitch & -- & -- & -- & -- & -- & -- \\
F4 & MCP & Yaw & -- & -- & -- & -- & -- & -- \\
F4 & MCP & \bf{Roll} & 0.7 & 0.7 & 1.2 & 1.0 & 2.5 & 2.5 \\
F4 & PIP & Pitch & -- & -- & -- & -- & \bf{5.7} & \bf{5.7} \\
F4 & \bf{DIP} & Pitch & \bf{5.7} & \bf{5.8} & \bf{7.5} & \bf{7.6} & \bf{25.9} & \bf{29.5} \\
\hline
\hline
F5 & MCP & Pitch & -- & -- & -- & -- & -- & -- \\
F5 & MCP & Yaw & -- & -- & -- & -- & -- & -- \\
F5 & MCP & \bf{Roll} & 2.8 & 2.8 & 2.8 & 3.5 & 1.3 & 1.6 \\
F5 & PIP & Pitch & -- & -- & -- & -- & -- & -- \\
F5 & \bf{DIP} & Pitch & \bf{20.2} & \bf{8.3} & \bf{21.2} & \bf{8.5} & 2.8 & \bf{12.1} \\
\hline
\end{tabular}
\caption{Maximum joint constraint violations observed during transient gesture execution under soft-constraint dynamics in the MuJoCo simulation environment.
For each finger, joint, and rotational degree of freedom, the table reports the maximum deviation (in degrees) of the achieved joint angle from its predefined feasible range during the control transients to six target gestures (G1--G6).
Entries marked with ``--'' indicate that no constraint violation occurred for the corresponding joint and gesture.
Boldface values highlight particularly large violations ($>5^\circ$) and identify the most critical joints affected during the transients.
The reported violations arise within MuJoCo’s soft-constraint formulation, in which joint limits are enforced through compliant penalty-based forces rather than hard constraints, allowing temporary excursions beyond the feasible domain during dynamic motion.
As an illustrative example, if the feasible range of the F5 DIP pitch angle is $[-2^\circ, 30^\circ]$ and the joint temporarily reaches $45^\circ$, the reported constraint violation is $15^\circ$. In this case study, we used PI feedback with feedforward compensation.}

\label{table:ConViolation}
\end{table}

\section{Conclusion}

This paper addressed the challenge of developing kinematic models to estimate the posture of the Anatomically Correct Biomechatronic Hand model \cite{2019_Tasi.etal} from the measured tendon tension and displacement. The paper highlighted that estimating the joint angles in such complex hand model is challenging due to the increased degrees of freedom in the bone structure and the complexity of the tendon length computations.

Although the solution of the nonlinear equations are possible with the efficient gradient-based solvers, the structural controllability of the developed in silico model should be addressed in the future.
This is especially true in the case of hindered movements, when unexpected physical constraint occurs. The uncertainty - inaccuracy experienced along axis Roll suspected to be caused by the given structural properties of the tendon system.

In summary, this article presents a novel approach for joint posture estimation in high-degree-of-freedom anthropomorphic robotic hand models and offers valuable insights into the complexities and challenges associated with achieving dexterity and control in such systems. The proposed methodology and the DH description provide a foundation for further research and advancements in the field of robotic hand manipulation and control.

\end{document}